%% file: neurips_2026.tex
\documentclass{article}
\usepackage[toc,page,header]{appendix}
\usepackage{minitoc}
\usepackage{graphicx}
\usepackage{amsmath}
\usepackage{subcaption}
\usepackage{adjustbox}
\usepackage{tcolorbox}
\usepackage{amsthm}
\newtheorem{theorem}{Theorem}
\usepackage{etoolbox}

\newcommand{\notocsection}[1]{%
  \begingroup
  \let\addcontentsline\relax
  \section{#1}
  \endgroup
}

\usepackage[main, preprint]{neurips_2026}

\usepackage[utf8]{inputenc} 
\usepackage[T1]{fontenc}    
\usepackage{hyperref}       
\usepackage{url}            
\usepackage{booktabs}       
\usepackage{amsfonts}       
\usepackage{nicefrac}       
\usepackage{microtype}      
\usepackage{xcolor}         

\title{Understanding Sample Efficiency in Predictive Coding}

\author{
Gaspard Oliviers\thanks{These authors contributed equally to this work.} \quad
Elene Lominadze\footnotemark[1] \quad
Rafal Bogacz \\
Nuffield Department of Clinical Neurosciences, University of Oxford, United Kingdom\\
MRC Centre of Research Excellence in Restorative Neural Dynamics, United Kingdom  \\
\texttt{Correspondence to: rafal.bogacz@ndcn.ox.ac.uk}
}

\begin{document}

\doparttoc 
\faketableofcontents 

\maketitle

\begin{abstract}
  Predictive Coding (PC) is an influential account of cortical learning. Much of recent work has focused on comparing PC to Backpropagation (BP) to find whether PC offers any advantages. Small scale experiments show that PC enables learning that is more sample efficient and effective in many contexts, though a thorough theoretical understanding of the phenomena remains elusive. To address this, we quantify the efficiency of learning in BP and PC through a metric called ``target alignment'', which measures how closely the change in the output of the network is aligned to the output prediction error. We then derive and empirically validate analytical expressions for target alignment in Deep Linear Networks. We show that learning in PC is more efficient than BP, which is especially pronounced in deep, narrow and pre-trained networks. We also derive exact conditions for guaranteed optimal target alignment in PC and validate our findings through experiments. We study full training trajectories of linear and non-linear models, and find the predicted benefits of PC persist in practice even when some assumptions are violated. Overall, this work provides a mechanistic understanding of the higher learning efficiency observed for PC over BP in previous works, and can guide how PC should be parametrised to learn most effectively.
\end{abstract}



\section{Introduction}

\begin{figure}
    \centering
    \includegraphics[width=\textwidth]{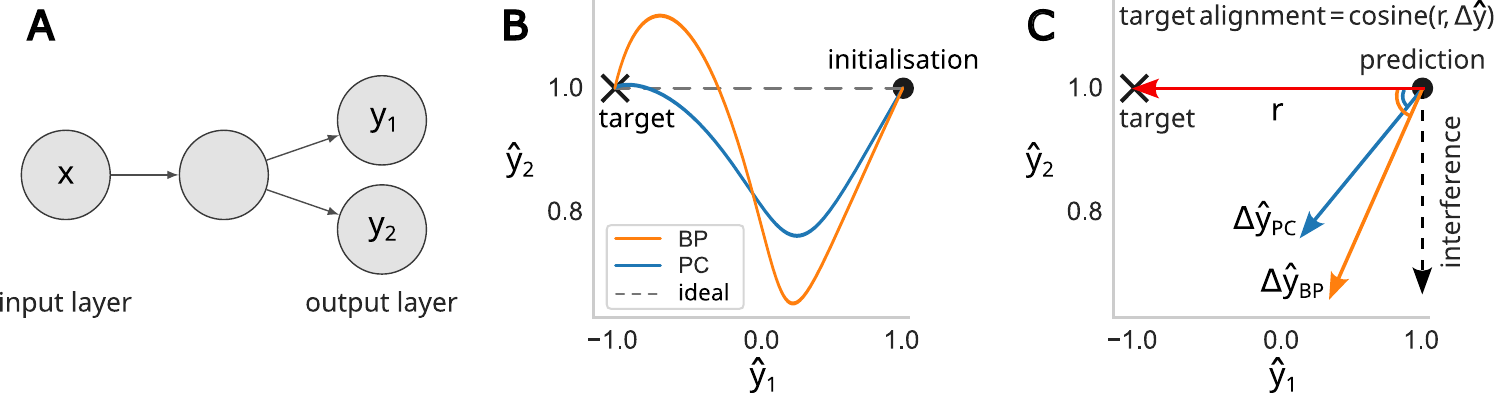}
    \caption{\textbf{Target alignment.} \textbf{(A)} Deep Linear Network (DLN) used in the toy model has one input neuron, one hidden neuron and two output neurons $y_1$ and $y_2$. All weights are initialised to be one. \textbf{(B)} Evolution of a DLN’s predictions $\hat y_i$ during training with BP and PC. The network receives an input of $x=1$ with a target output of $y=[-1,1]$. Training is carried out separately using PC and BP until the model output matches the target.
    The learning trajectory shown with the grey dashed line is ideal, as the model’s prediction changes only for incorrectly predicted output $\hat y_1$, remaining unchanged for  $\hat y_2$ that is already well predicted.
    PC changes its output more directly towards the target than BP. \textbf{(C)} Definition of target alignment, where $r = y - \hat y$, while blue and orange vectors show changes in output $\Delta \hat y$ after one weight update. Additionally a dashed arrow shows direction perpendicular to the output error $r$ labelled `interference' because change in model prediction, $\Delta \hat y$, that is partially aligned with it creates errors in directions the model already predicts correctly.}
    \label{fig:ch3_ta_n_lin_traj}
\end{figure}

Predictive coding (PC) offers an influential account of cortical learning based on minimising prediction errors through local weight updates \cite{rao1999predictive, friston2003learning,Bogacz2017}. PC differs markedly from backpropagation (BP), the dominant training paradigm in modern machine learning \cite{rumelhart1986learning, goodfellow2016deep, Krizhevsky2012}. Recent work \cite{song2024inferring} has suggested that PC is more sample efficient and outperforms BP on learning regimes that are both biologically relevant and particularly prone to catastrophic interference, such as learning from limited data, online parameter updates, and continual learning. It has been argued \cite{song2024inferring} that these advantages arise because PC parameter updates induce changes in the model’s predictions that are better aligned with the output error than those induced by BP (Figure \ref{fig:ch3_ta_n_lin_traj}A-B).
Changes in the model’s predictions that are misaligned with the output error are referred to in this paper as \emph{interference} because they introduce errors in directions the model already predicts accurately, so that incorporating new information interferes with preserving prior knowledge. The lack of interference can be quantified by \emph{target alignment}, which measures the similarity between the output error during training and subsequent changes in network's prediction (Figure \ref{fig:ch3_ta_n_lin_traj}C).
The more effective learning performance of PC over BP could therefore be explained by PC's higher target alignment. 

An additional perspective arises from the observation that optimal target alignment is equivalent to performing learning along natural gradients \cite{AmariNaturalGradients}. In this regime, the change in the model’s predictions is exactly aligned with the output error, yielding the steepest descent in output error \cite{AmariNaturalGradients}. Natural gradient methods are known to accelerate learning \cite{Hennequin2028NaturalGrad, Dongsung2020, martens2020optimizingneuralnetworkskroneckerfactored}, reduce sensitivity to parameterisation \cite{martens2020new}, and improve learning stability \cite{jnini2025dualnaturalgradientdescent, muller2023achievinghighaccuracypinns}. In deep linear networks (DLNs), they also render online and batch learning equivalent \cite{AmariNaturalGradients}. Interpreted through this lens, target alignment provides a measure of how closely learning dynamics approximate natural gradient descent, thereby indicating potential benefits of PC over BP by operating closer to the natural gradient regime.

However, a principled analytical explanation of when and why PC should outperform BP, and on what classes of problems, has remained elusive. Moreover, existing evidence of potential benefits of PC has largely been confined to small-scale models \cite{song2024inferring}. 

This paper provides an analytical account of the interference-reducing properties of PC. We show that, in DLNs, PC reduces the weight-dependent interference that arises in BP. As a consequence, PC exhibits improved target alignment, especially in regimes where model weights are poorly conditioned, including architectures with information bottlenecks and pre-trained models.  We derive closed-form expressions for the change in prediction induced by PC and BP updates in DLNs and compare their target alignment. We also characterise how PC’s update can be modified to guarantee interference-free learning for learning with individual samples and for a batch of inputs. We show the theoretical benefits offered by higher target alignment can lead to improved sample efficiency throughout longer training trajectories.

\subsection*{Summary of contributions}
\begin{itemize}
    \item We show theoretically that learning in PC has less interference than BP, which is especially pronounced in deep, narrow and pre-trained networks.    
    \item We demonstrate that PC with layer-specific learning rates guarantees interference-free single-sample learning, while PC with weight update rescaling factors enables interference-free batch learning.
    \item We show that the benefits predicted from the theory translate to longer training trajectories, making PC more sample efficient. We give preliminary evidence that these findings can be extended to different architectures and nonlinear networks.
\end{itemize}

\vspace{-2pt}
\section{Background and Related Work}
\label{gen_inst}

\subsection{Deep linear networks}
Deep linear networks (DLNs) are fully connected linear networks with one or more hidden layers. DLNs have been extensively used to analytically describe the learning dynamics of BP \cite{saxe2014exact, NIPS2016_f2fc9902, Hennequin2028NaturalGrad, domine2024lazy} and PC \cite{millidge2022theoreticalframeworkinferencelearning, InnocentiSadles}. Despite learning linear models, DLNs exhibit non-linear learning dynamics. These learning dynamics have proved useful for understanding non-linear networks\cite{saxe2014exact}. 
Similarly, DLNs have been used to analyse PC. For instance, it has been shown that PC's equilibrium activity interpolates between the feedforward dynamics of BP and backwards-propagated targets, such as those used in target propagation \cite{millidge2022theoreticalframeworkinferencelearning}. 

For a hierarchical model with $L$ layers, a DLNs learns with BP by minimising a standard squared error energy function (which we also rewrite as a dot product for easy comparison with PC below):
\begin{equation}
    \mathcal{L}_{BP}(x, y, W) = \tfrac{1}{2}||y - W_{L:1}x||^2_2
    = \tfrac{1}{2}(y - W_{L:1}x)^\top (y-W_{L:1}x)
    \label{eq:ch3_loss_bp}
\end{equation}
where $x$ is an input activity, $y$ is a target output activity, the parameter $W$ includes the weight matrices $W_l$ mapping layer $l-1$ to layer $l$, and $W_{L:1}$ denotes the product of weights from the input to the output layer, $W_{L:1} = W_L...W_1$.

\subsection{Learning with predictive coding}
DLNs can also learn using PC. In a PC model with $L$ layers, the activity in each layer is denoted by $x_l$, with $x_0$ the input layer and $x_L$ the output layer. During training the input layer is clamped to $x_0=x$, and the output layer $x_L$ is fixed to the target $x_L=y$. PC trains a linear DLN by minimising the energy function:
\begin{equation}
    E(x_0,...,x_L, W) = \sum_{l=1}^L \tfrac{1}{2}||x_l - W_{l}x_{l-1}||^2_2.
    \label{eq:ch3/pc_base}
\end{equation}
This energy function is minimised through two steps. First, the neural activity of the network is updated to minimise the energy until convergence, and we refer to this process as \emph{inference}. Then, the weights are updated using one gradient step on the energy function, evaluated at the activity minimising energy found during inference.

For a DLN, PC's inference and learning dynamics are analytically tractable. 
The neural activity in equilibrium of inference $x_l^*$ can be calculated analytically
\cite{millidge2022theoreticalframeworkinferencelearning}.
Substituting the equilibrium activity in hidden layers ($x_l=x_l^*$ for $1<l<L$) into the energy function of equation \ref{eq:ch3/pc_base}, allows expressing the energy of PC as a function of just input, target and weights
\cite{InnocentiSadles}: 
\begin{equation}
\label{eq:rescaledEnergy}
    E(x, y, W) = \tfrac{1}{2}(y - W_{L:1}x)^\top S^{-1}(y-W_{L:1}x). 
\end{equation}
This energy has analogous forms as for BP (equation \ref{eq:ch3_loss_bp}) differing only in additional term $S$ given by:
\begin{equation}
    S = I + \sum_{l=1}^{L-1} W_{L:l+1}W_{L:l+1}^\top .
    \label{ch3/eq:S}
\end{equation}
Taking a gradient of equation \ref{eq:rescaledEnergy} will allow us to find analytically the change of weights in PC (which in the model is normally computed through inference and subsequent local weight updates). 

\subsection{Target alignment of learning}
Learning updates that follow the shortest path between the current prediction and the target reduce interference \cite{song2024inferring, kao2021naturalcontinuallearningsuccess} and lead to faster convergence \cite{AmariNaturalGradients,fast_converge2}. Neither BP nor PC follows this shortest path for a simple DLN with architecture shown in Figure \ref{fig:ch3_ta_n_lin_traj}A. 
Figure \ref{fig:ch3_ta_n_lin_traj}B illustrates how the output prediction of this DLN changes during training. It shows that PC learning trajectories are more closely aligned with the ideal path than those of BP. This behaviour has been observed in both small linear models and deeper non-linear architectures \cite{song2024inferring}.

The extent to which a learning rule follows the ideal update direction can be quantified using target alignment. Target alignment measures how strongly the change in a model’s prediction induced by a weight update is aligned with its prediction error, using cosine similarity \cite{song2024inferring}:
\begin{equation}
    \text{Target alignment} = cos(\angle(r, \Delta \hat y)) = \frac{r^\top  \Delta \hat y}{\|r\|\ \|\Delta \hat y\|}
    \label{eq:ch3_target_alignmet_eq}
\end{equation}
Here,  $r$ denotes the output prediction error $r = y - W_{L:1}x$, $\hat y= W_{L:1}x$ is the feedforward prediction, and $\Delta\hat y$ represents the change in prediction resulting from a weight update. For DLNs, $\Delta\hat y = (W_{L:1}^{\text{post}} -W_{L:1}^{\text{pre}}) x$, with pre and post indicating the weights' state before and after the update respectively.
A target alignment of one indicates perfect alignment, zero indicates orthogonality, and minus one indicates opposite directions. 
Definition of target alignment for PC and BP is illustrated in Figure \ref{fig:ch3_ta_n_lin_traj}C.

The reasons why PC exhibits higher target alignment than BP remain unclear. It has been shown that PC inference combines BP-like feedforward activity with signals resembling target propagation \cite{millidge2022theoreticalframeworkinferencelearning}, and that models trained with target propagation achieve optimal target alignment \cite{meulemans2020theoreticalframeworktargetpropagation, song2024inferring}. From this perspective, PC may attain higher target alignment than BP by effectively integrating BP with target-propagation-like updates \cite{song2024inferring}. However, this explanation remains empirical and lacks analytical proof.

Alternative learning algorithms with higher target alignment than BP have been studied extensively in machine learning. In particular, natural gradient methods \cite{AmariNaturalGradients} guarantee maximally aligned learning updates by rescaling BP parameter updates with the inverse Fisher information matrix.
By accounting for interactions between parameters in their effect on the model output, the Fisher matrix reconditions the gradient to produce updates that achieve optimal target alignment and maximise the reduction of the objective function per update. Natural gradient methods have also been investigated in the context of DLNs \cite{Hennequin2028NaturalGrad}, 
but the relevance of natural gradients to large-scale deep learning and to neuroscience remains limited.

\section{Results}
\label{headings}

\subsection{PC is more sample efficient for training deep, narrow, and pre-trained networks than BP}

To understand when PC yields higher target alignment than BP, we first derive the change in output prediction of a DLN trained with these algorithms. Specifically, we analyse the alignment between changes in the network’s prediction and the training output residual, $r = y - \hat{y}$.
We show that, in BP, the change in prediction is given by a sum of terms in which $r$ is pre-multiplied by a weight matrix. This matrix distorts the direction of $r$, thereby introducing interference. In contrast, PC reduces this interference. As a consequence, PC exhibits improved target alignment, especially in deep models that are either narrow in width or with poorly conditioned weights as could be observed in a pre-trained model. We also show that PC's higher target alignment leads to more sample efficient learning than BP. These results are derived for models with an MSE output loss. However, Appendix \ref{app:ce} generalises the result to arbitrary loss functions.

\subsubsection{Interference in BP}

In BP, the weight updates follow the steepest descent of the model's objective function with respect to the weights. 
Although the weight updates occur in discrete steps, we analyse the corresponding continuous-time limit, which holds as long as the learning rate is sufficiently small. This approach aligns with prior studies on the weight dynamics of DLNs under BP \cite{saxe2014exact, Hennequin2028NaturalGrad} and PC \cite{InnocentiSadles}.
The weight dynamics for an input $x$ and output $y$ are given by:
\begin{equation*}
    \frac{\partial W_l}{\partial t} = -\nabla_{W_l}\mathcal{L}_{BP} =  W_{L:l+1}^\top (y - W_{L:1}x) (W_{l-1:1}x)^\top =  W_{L:l+1}^\top r (W_{l-1:1}x)^\top.
\end{equation*}
The change in prediction resulting from the weight dynamics can then be calculated by differentiating using the product rule, as follows:
\begin{align}
    \frac{\partial \hat y}{\partial t}    &=   \sum_{l=1}^L W_{L:l+1}\frac{\partial W_l}{\partial t} W_{l-1:1}x =   \sum_{l=1}^L W_{L:l+1} W_{L:l+1}^\top r \hat x_{l-1}^\top  \hat x_{l-1}
    \label{eq:ch3_BP_deltapred}
\end{align}
where the feed forward activity at layer $l$ equal to $\hat x_{l}=W_{l:1}x$, and 
$W_{L:L+1}$ and $W_{0:1}$ are equal to the identity matrix. This result is equivalent to previous derivations for DLNs \cite{Hennequin2028NaturalGrad}.


The derived expression for the change in prediction under BP reveals a single source of interference. In the ideal case, the change in prediction would equal the target residual $r$. However, the actual update is given by a sum in which each term consists of $r$ pre-multiplied by $W_{L:l-1}W_{L:l-1}^\top$ and post-multiplied by $\hat x_{l-1}^\top\hat x_{l-1}$.
The matrix $W_{L:l-1}W_{L:l-1}^\top$ acts directly on $r$, thereby distorting its direction and introducing interference. In contrast, the factor $\hat x_{l-1}^\top\hat x_{l-1}$ is a scalar rather than a matrix. Therefore, it only rescales the update magnitude and does not contribute to interference.

Interference is therefore expected to be large when $W_{L:l-1}W_{L:l-1}^\top$ deviates substantially from the identity matrix. This occurs when $W_{L:l-1}$ is far from orthogonal, which can arise from two sources:
(i) architecture: deeper networks introduce more interacting terms, while narrower layers are prone to rank deficiency, increasing interference. (ii) training: weights may develop correlated features and amplify certain feature directions while suppressing others, leading to poor conditioning.


\subsubsection{Interference in PC}
\label{sec:interferenceinPC}

Interference in a network trained with PC differs from that in BP due to an additional scaling matrix $S$ in equation \ref{eq:rescaledEnergy}. This additional term counteracts distortions induced by the weights, attenuating interference. 
We derive the change in prediction $\frac{\partial \hat y}{\partial t}$ of a DLN trained with PC's objective function $E$ given in equation (\ref{eq:rescaledEnergy}). 
We show in Appendix \ref{s:ch3_deriV_pc_update} that weight dynamics of PC are given by:
\begin{equation}\label{ch3/weightupdates}
    \frac{\partial W_l}{\partial t} = - \nabla_{W_l}{E} = W_{L:l+1}^\top\ S^{-1} r\ {x}_{l-1}^{*\top}
\end{equation}

The change in prediction resulting from PC's weight dynamics can be calculated by differentiating using the product rule, as follows:
\begin{align}
    \frac{\partial \hat y}{\partial t} 
    &=   \sum_{l=1}^L W_{L:l+1}\frac{\partial W_l}{\partial t} W_{l-1:1}x =   \sum_{l=1}^L W_{L:l+1} W_{L:l+1}^\top\ S^{-1} r\ {x}_{l-1}^{*\top} \hat x_{l-1}
    \label{eq:ch3_PC_deltapred}
\end{align}

Crucially, this expression differs from BP (equation \ref{eq:ch3_BP_deltapred}) through the scaling matrix $S^{-1}$. $S$ was defined in equation~(\ref{ch3/eq:S})
and can be written more compactly using notation $W_{L:L+1}=I$ introduced earlier as:
\begin{align*}
    S = \sum_{l=1}^{L} W_{L:l+1}W_{L:l+1}^\top. 
\end{align*}
The structure of $S^{-1}$ enables cancellation with the weight-dependent terms $W_{L:l+1} W_{L:l+1}^\top$ in the prediction change (equation \ref{eq:ch3_PC_deltapred}), thereby attenuating interference. When the scalar factor $x_{l-1}^{*\top}\hat{x}_{l-1}$ is constant across layers, interference matrices cancel completely. Under this assumption $x_{l-1}^{*\top}\hat{x}_{l-1} = c$ for all $l$, the PC target alignment simplifies to:
\begin{align}
    \frac{\partial \hat y}{\partial t}
    & = \sum_{l=1}^L W_{L:l+1} W_{L:l+1}^\top S^{-1} r \, x_{l-1}^{*\top} \hat x_{l-1} = \underbrace{\sum_{l=1}^L W_{L:l+1} W_{L:l+1}^\top}_{S} S^{-1} r \, c = r\ c,
    \label{eq:perfect}
\end{align}

This reveals PC's intrinsic interference-reduction mechanism via $S^{-1}$, independent of network architecture or training stage. While $x_{l-1}^{*\top}\hat{x}_{l-1}$ is not strictly constant across layers in practice, the scaling matrix $S^{-1}$ substantially reduces interference in PC over BP whenever these quantities remain reasonably consistent.


\begin{tcolorbox}[colback=blue!3!white, colframe=blue!12!white, coltext=black, coltitle=black, title=\textbf{Theoretical Takeaway: When will PC have significantly less interference than BP?}]
    PC has higher target alignment than BP, assuming pre- and post-inference activity norms in PC are approximately constant across layers. 
    The difference will be significantly different when BP's target alignment is low due to:
    \begin{itemize}
        \item \textbf{Architecture.} BP interference increases in \textbf{deep} and \textbf{narrow} networks, whereas PC interference is less affected by architecture.
        \item \textbf{Training.} As weight conditioning deteriorates during training, BP interference increases, while PC interference is much less affected.
    \end{itemize}
\end{tcolorbox}

\subsubsection{Higher target alignment in PC leads to more sample-efficient learning than BP}
\label{sec:HigherTAinPC}




We experimentally validate our theoretical predictions that PC has higher target alignment than BP for single weight updates in randomly initialised networks. We also show that PC's higher target alignment leads to more sample efficient learning. 

We first evaluate target alignment in randomly initialised DLNs after a single weight update using either BP or PC. We vary network depth and width.
We compare two weight initialisation schemes: Kaiming uniform initialisation, used as Pytorch's default initialisation for linear layers, and 
Norm-preserving initialisation, which aims to preserve the norm of activity ($||x_l||^2_2$) across all layers in expectation. Kaiming initialisation causes the activity norm of a network to decay rapidly with depth, whereas Norm-preserving initialisations preserve stable activation norms (see Appendix \ref{sec:Norm-preservingInit}), and thus brings the model closer to satisfying the assumption $x_{l-1}^{*\top}\hat{x}_{l-1} = c$, enabling perfect target alignment (equation \ref{eq:perfect}).
Unless stated otherwise, we consider a DLN with input, hidden, and output widths all set to 512, a single hidden layer. Inputs and targets are sampled independently from standard normal distributions, the batch size is one, and results are averaged over ten random seeds.

\begin{figure}[t]
    \centering
    \includegraphics[width=1\textwidth]{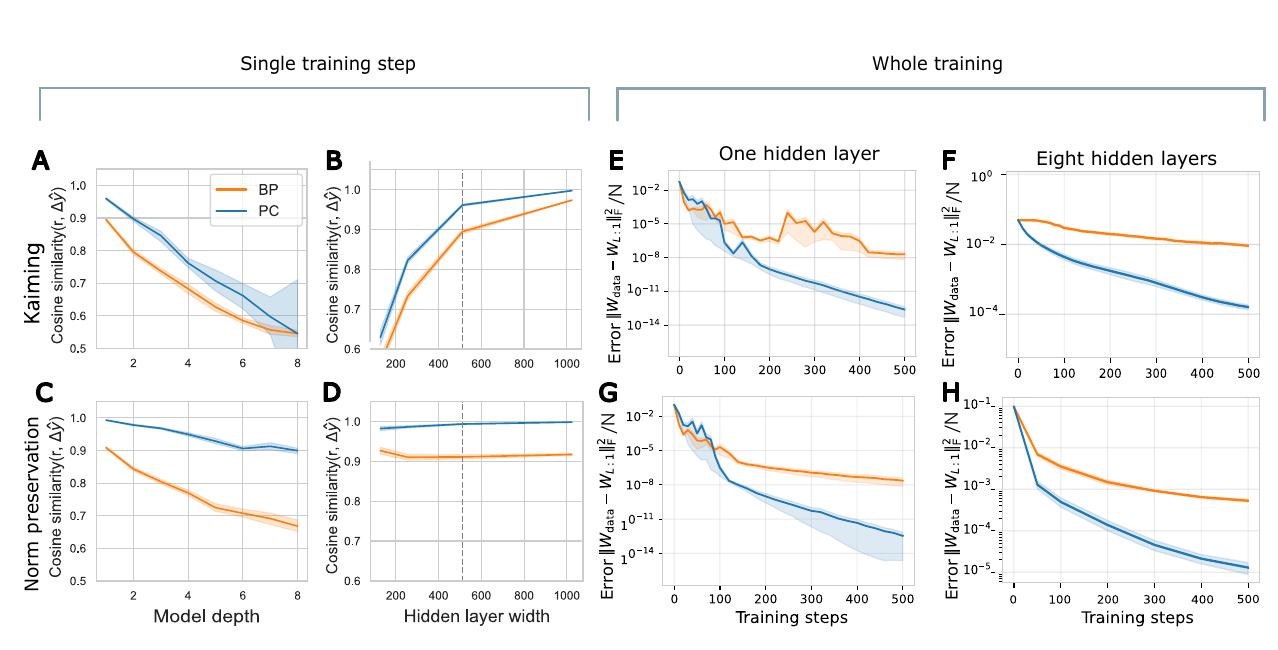}
    \caption{\textbf{Target Alignment in Predictive Coding (PC) and Backpropagation (BP).} \textbf{(A-D)} Target alignment of BP (orange) and PC (blue) after one training step for a deep linear network with different architectures and initialisations. The dashed grey lines in panels \textbf{B} and \textbf{D} denote square networks. \textbf{(E, G)} Training dynamics of linear BP and PC under different initialisations and a batch size of $64$. The models are square with $20$ units per layer and one hidden layer. They are initialised identically and are exposed to identical batches at each training step. \textbf{(F, H)} Identical setup to experiments in panels \textbf{E} and \textbf{G} but for networks with eight hidden layers. The high variability of PC's target alignment for Kaiming initialisation is investigated in Appendices \ref{appendix:postmult} and \ref{app:isolateS}.
    }
    \label{fig:fig2}
\end{figure}

Figure \ref{fig:fig2}A-D shows that PC achieves better target alignment than BP under both Kaiming and Norm-preserving initialisations. These results indicate that PC’s interference-reducing mechanism remains effective even when the activity norm is not constant across layers. Norm-preserving initialisation attains near-perfect alignment across settings. Although alignment decreases slightly with increasing depth, it remains largely insensitive to changes in network width. In contrast, Kaiming initialisation results in poorer overall alignment. Under Kaiming initialisation, PC alignment deteriorates markedly as depth increases and exhibits greater variability across model configurations.


We further evaluate how PC and BP perform over whole training trajectories rather than one training step. To do this, we set up a simple synthetic regression task, whereby the goal of the network is to learn a mapping $W_{data}\sim \mathcal{N}(0,1/d_{in})$ where $d_{in}$ is the input dimension and $W_{data} \in \mathbb{R}^{d_{out}\times d_{in}}$ ($\mathcal{N}(0,\Sigma)$ denotes a normal distribution with mean 0 and variance $\Sigma$, see Appendix \ref{Appendix:TrainingData} for further details). During training, we measure the quantity $\frac{1}{N}||W_{data}-W_{L:1}||_F^2$ where $N=d_{out}\times d_{in}$. For the experiments labelled ``Whole training'' in Figure \ref{fig:fig2}, we choose a square neural network with $20$ input, hidden, and output units similar to \cite{Hennequin2028NaturalGrad}. We vary the depth of the network between $1$ and $8$ hidden layers, as well as varying the initialisation. By default, we set the batch size to $64$ unless stated otherwise. We perform learning rate sweeps for both BP and PC, identifying optimal learning rates for each by sampling $100$ points within the region $10^{-3.5}-10^{0.3}$. We average our results over ten random seeds.


Figure \ref{fig:fig2}E-H shows that for both initialisations, Norm-preserving and Kaiming, PC is more sample efficient while learning over longer training trajectories. Panels E and G are consistent with the results from panels A and C, showing that for a square network with one hidden layer, PC outperforms BP. For eight hidden layer networks depicted in panels F and H, PC with Norm-preserving initialisation clearly outperforms all alternatives considered for the given architecture. The expected benefits also translate when varying widths as demonstrated in Appendix \ref{sec:AdditionalExperimentalData}. Moreover,  the results are confirmed on CIFAR-10, CIFAR-100 and STL-10 in Appendix \ref{app:ae}.


\subsection{Rescaling learning rate for online learning}
\label{sec:online_learning}

Although our previous results show that PC exhibits better alignment than BP, its standard formulation does not achieve optimal target alignment. 
Introducing layer-specific learning rates, ensures that PC achieves optimal alignment for linear models trained on a single sample, irrespective of neural activities, model weights or model architecture.
\begin{tcolorbox}[width=\linewidth, sharp corners=all, colback=white!95!black, colframe=white!95!black]
    \begin{theorem}[PC with adaptive learning rate]
    \label{theorem1}
    For layer-specific adaptive learning rates with $\alpha_l$ set to the scalar $(x_{l-1}^{*\top}\hat{x}_{l-1})^{-1}$, PC achieves optimal target alignment when learning on an individual input-output pair.  For PC's weight update 
    \begin{equation*}
    \frac{\partial W_l}{\partial t} = -\alpha_l \nabla_{W_l}{E},
    \quad \text{for } l = 1, \dots, L-1
\end{equation*}
    the change in model prediction is given by:
    \begin{equation*}
        \frac{\partial \hat y}{\partial t} = r        
    \end{equation*}
    \end{theorem}
\end{tcolorbox}
This theorem is proven in Appendix \ref{appendix:thorem1}.

To assess whether layer-specific learning rates allow PC to achieve perfect alignment, we measure target alignment in a DLN trained with and without scaling under both BP and PC. 
For BP, we scale the BP weight updates by the inverse of corresponding scaling factors, $\alpha_l=(\hat x_{l-1}^\top\hat x_{l-1})^{-1}$, ensuring a fair comparison.
We vary the conditioning of the model weights to illustrate the effect of training discussed in section \ref{sec:interferenceinPC}.
The conditioning is quantified using the condition number (see details in Appendix \ref{sec:conditioning_metric}), which is the ratio of the largest to the smallest singular value. Larger condition numbers indicate poorer conditioning. 
All other experimental settings are set to the default values of the previous experiments, and models were initialised using Kaiming uniform initialisation.

\begin{figure}[t]
    \centering
    \includegraphics[width=1\textwidth]{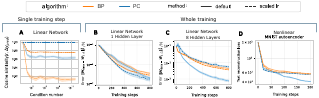}
    \caption{\textbf{Rescaling learning rate for online learning.} \textbf{(A)} Target Alignment as a function of the condition number of the weight matrix for four models, BP, PC and their rescaled counterparts. \textbf{(B)} Training dynamics of a square linear network with $20$ units per layer. \textbf{(C)} Training of the same square network with $8$ hidden layers. Curves are averaged over 10 runs. \textbf{(D)} Nonlinear autoencoder with 3 hidden layers and trained on online MNIST reconstruction task. Results averaged over $3$ seeds.}
    \label{fig:fig3}    
\end{figure}

Figure \ref{fig:fig3}A confirms that introducing layer-specific learning rates ensures that PC always learns with the maximal target alignment of one. 
BP benefits from layer-specific scaling factors as well, but does not achieve maximal alignment or independence from architecture. 
Increasing the condition number of the weight matrices reduces target alignment for both PC and BP, except for PC with activity-dependent learning rates. This behaviour is expected, as a higher condition number corresponds to stronger interference from the weight product $W_{L:l+1}W_{L:l+1}^\top$.


In panels B and C of Figure \ref{fig:fig3}, we again initialised DLNs with Norm-preservation and a square architecture with $1$ hidden layer in panel B and $8$ hidden layers in panel C.  We tested all four models on the online synthetic regression task and carried out extensive learning rate sweeps over the range $10^{-3.5}-10^{-0.04}$ (see Appendix \ref{sec:1LayerVaryingWidth}). 
Figure \ref{fig:fig3}B shows the training dynamics of the square network. In the online setting, while PC and scaled PC still outperform BP and scaled BP, the scaling provides little benefit for sample efficiency at the beginning of training. This also holds for deeper networks displayed in Figure \ref{fig:fig3}C. To understand why this happens, we draw an analogy to natural gradients, which is what the weight updates for rescaled PC reduce to for larger batch sizes. Estimating the Fisher Information Matrix with a single data point, as we do for online learning, diminishes its geometric advantages. However, some benefits can still be observed as the number of training steps becomes large. Appendix \ref{sec:AdditionalExperimentalData} shows experimental results as the width is varied between $15, 20$ and $40$ hidden units for a DLN with $1$ hidden layer. 


Figure \ref{fig:fig3}C shows a small nonlinear autoencoder with $3$ hidden layers trained on MNIST \cite{lecun2010mnist} (see Appendix \ref{sec:TrainingforMultipleSteps} for architecture and training details). Again, rescaling provides limited benefit for batch size $1$.
We expect the performance of PC and rescaled PC to greatly benefit from batch training due to the emerging similarity between rescaled PC and natural gradients as the batch size grows.

\vspace{-.3cm}

\subsection{Weight update rescaling for batch training}
\label{sec:WeightRescalingBatchLearning}


Training on a mini-batch introduces additional sources of interference in both BP and PC. This interference depends on the norm of layer activities within and across samples. This result is derived in Appendix \ref{appendix_batch_train} by calculating the change in predictions of a DLN when training on a mini-batch of data for BP and PC.

Standard PC does not have a mechanism to reduce inter-sample interference. However, PC can maximise target alignment during multi-sample learning by multiplying its weight updates by a matrix that removes correlations between sample-specific updates.
\vspace{-.08cm}
\begin{tcolorbox}
[width=\linewidth, sharp corners=all, colback=white!95!black, colframe=white!95!black]
    \begin{theorem}[PC with weight update rescaling]
    \label{theorem2}
    For the layer-specific weight update rescaling factors 
    $A_l$ equal to $(\frac{1}{B}\sum_{b=1}^B[\hat x_{l-1,b} \  x_{l-1,b}^{*\top} ])^{+}$, for the pseudo inverse operation $.^+$ when the batch size $B$ is smaller or equal than the smallest width of activity layers $0$ to $L-1$, PC achieves optimal target alignment for batch learning. For PC's weight update
    \begin{equation*}
        \frac{\partial W_l}{\partial t} = -\frac{1}{B}\sum_{b=1}^B\nabla_{W_l}{E}\ A_l,
    \quad \text{for } l = 1, \dots, L
    \end{equation*}
    the change in model prediction is given by:
    \begin{equation*}
        \frac{\partial \hat y_b}{\partial t} =  r_b,
    \quad \text{for } b = 1, \dots, B
    \end{equation*}
    \end{theorem}
\end{tcolorbox}
This theorem is proven in Appendix \ref{appendix_theorem2}. 
While the leaning rule in theorem \ref{theorem2} is not local,
factor $A_l$ is closely linked to the precision of neural activities, which can be computed by PC's neural circuits \cite{Bogacz2017}. 

We repeat the target alignment experiments with a single weight update while varying the batch size.
For BP, we employ a scaling matrix $A_l = (\frac{1}{B}\sum_{b=1}^B[\hat x_{l-1,b} \hat x_{l-1,b}^{\top}])^{-1}$. To ensure numerical stability, we employ Moore–Penrose pseudoinverses.

\begin{figure}[!t]
    \centering
    \includegraphics[width=1\textwidth]{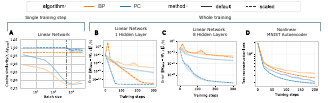}
    \caption{\textbf{Weight update rescaling for batch training} \textbf{(A)} Target Alignment as a function of the batch size for four models, BP (solid orange), PC (solid blue) and their rescaled counterparts (dashed). \textbf{(B)} Training dynamics of the models trained with a batch size of $64$ for a $1$ hidden layer square linear network with 20 units per layer, as well as its deeper counterpart with $8$ hidden layers \textbf{(C)}. \textbf{(D)} Nonlinear autoencoder with 3 hidden layers and ReLU activations trained on MNIST reconstruction task with batch size 64. Runs averaged over $10$ seeds for MLPs and $3$ for Autoencoders.}
    \label{fig:fig4}
\end{figure}

Introducing layer-wise scaling matrices enables PC to achieve optimal target alignment for batch training (Figure \ref{fig:fig4}A).
The target alignment of both PC and BP in their standard form decreases with increasing batch size.
PC with weight update scaling achieves optimal target alignment independently of batch size. 
BP also exhibits improved alignment with scaling matrices, but still fails to reach optimal alignment and consistently underperforms PC.


For whole training experiments, we keep the protocol outlined in section \ref{sec:online_learning} but change the batch size from $1$ to $64$ and increase the range used to find the optimal learning rate to $10^{-3.5}-10^{0.4}$. 
In order to ensure numerical stability when inverting covariance matrices, we
apply a small diagonal regularisation of the form $\tilde{A_l} = A_l + \varepsilon I$, $\varepsilon > 0$ is a scalar regularisation parameter. 
We select $\varepsilon$ based on the spectrum
of $\Sigma$ such that $\varepsilon = \max\bigl(0, \alpha \lambda_{\max} - \lambda_{\min}\bigr)$. 
$\alpha$ is set to $10^{-5}$ for networks with $1$ hidden layer and to $10^{-4}$ for networks with $8$ hidden layers. In panels B and C of Figure \ref{fig:fig4}, the default PC and BP implementations made fast initial progress but 
slowed down soon, 
scaled PC 
reached the minimum achievable error with much less training steps. Figure \ref{fig:fig4}D suggests that the advantages seen in linear networks extend to nonlinear Autoencoders. Rescaled PC reached the minimum test error fastest, followed by default PC, rescaled BP and BP.

\vspace{-.3cm}
\subsection{Target alignment with a finite number of inference steps}

The analysis so far has assumed equilibrated energy, though in practice equilibrium is often not reached due to a finite number of inference steps taken. Previous theorems can be extended to show that, if parameter updates are appropriately rescaled, perfect target alignment can be obtained regardless of the number of inference steps for online and batch learning.\\ 

\vspace{-.5cm}
\begin{tcolorbox}
[width=\linewidth, sharp corners=all, colback=white!95!black, colframe=white!95!black]
    \begin{theorem}[PC rescaling with finite number of inference steps]
    \label{theorem:finite_T_alignment}
    Let $T$ denote the number of inference steps and let $x^{(T)}$ be the activity of neurons after $T$ inference steps. For the layer-specific weight update rescaling factors 
    $A_l^{(T)}$ equal to $(\frac{1}{B}\sum_{b=1}^B[\hat x_{l-1,b} \hat x_{l-1,b}^{(T)\top} ])^{+}$, for the pseudo inverse operation $.^+$ when the batch size $B$ is smaller than or equal to the smallest width of layers $1$ to $L-1$, PC achieves optimal target alignment for batch learning. For PC's weight update
    \begin{equation*}
        \frac{\partial W_l}{\partial t} = -\frac{1}{B}\sum_{b=1}^B\nabla_{W_l}{E_b^{(T)}}\ A_l^{T},
    \quad \text{for } l = 1, \dots, L-1
    \end{equation*}
    the change in model prediction is given by:
    \begin{equation*}
        \frac{\partial \hat y_b}{\partial t} =  r_b,
    \quad \text{for } b = 1, \dots, B.
    \end{equation*}

    For online learning ($B=1$), a scalar
rescaling of  $\alpha_l^{(T)}
    =
    \frac{1}{
    x_{l-1}^{(T)\top}\hat{x}_{l-1}
    }$ is sufficient. 
    \end{theorem}
\end{tcolorbox}

Theorem \ref{theorem:finite_T_alignment} is proven in Appendix \ref{proof:3}. The form of the rescaling is very similar to those in theorems \ref{theorem1} and \ref{theorem2}, but equilibrated activities $x^*$ are replaced by $x^{(T)}$ (activities after $T$ inference steps). 

\begin{figure}[h]
    \centering
    \includegraphics[width=1\textwidth]{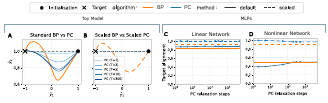}
    \caption{\textbf{Target alignment with and without rescaling as a function of inference steps} \textbf{(A)} Evolution of DLN's predictions $\hat{y}$ during training with BP and PC as a function of the number of inference steps $T$ (legend displayed in panel B) for the toy example discussed in figure \ref{fig:ch3_ta_n_lin_traj}. Grey horizontal dashed line corresponds to the trajectory with perfect target alignment. \textbf{(B)} Evolution of predictions $\hat{y}$ with rescaled weight updates following theorem \ref{theorem:finite_T_alignment}. All models with finite $T$ still follow the optimal trajectory with appropriate rescaling. \textbf{(C)} Target alignment in a deeper linear network after one weight update with and without rescaling as a function of the number of inference steps. \textbf{(D)} MLP with ReLU nonlinearities with and without rescaling as a function of inference steps.}
    \label{fig:fig5}
\end{figure}


To further investigate the implications of theorem \ref{theorem:finite_T_alignment}, we numerically simulate target alignment after one weight update (with and without rescaling) as a function of the number of inference steps in a toy model from Figure \ref{fig:ch3_ta_n_lin_traj}, a larger linear MLP, and a non-linear MLP. The training task for the toy model is identical to the setup in Figure \ref{fig:ch3_ta_n_lin_traj}, while the larger MLPs in Figures \ref{fig:fig5}C-D are trained on the same synthetic dataset as before (batch size of $128$) and have $512$ neurons per layer with $1$ hidden layer. We use a rate of $10^{-2}$ for parameter and activity updates. Figure \ref{fig:fig5}A shows that target alignment does not necessarily improve with increasing number of inference steps in standard PC without rescaling. Because of the setup of the toy problem (see Figure \ref{fig:ch3_ta_n_lin_traj}A), small $T$ means the shared parameters are minimally altered leading to less interference and therefore higher target alignment, however, this corresponds to a lazy learning regime where 
learning largely occurs in the final layer. Regardless of $T$, however, Figure \ref{fig:fig5}B confirms that introducing layer-specific rescaling ensures optimal target alignment in PC as long as the assumptions of theorem \ref{theorem:finite_T_alignment} are satisfied. This result holds in larger linear MLPs (Figure \ref{fig:fig5}C), and the benefits of rescaling may also translate to non-linear networks as demonstrated in Figure \ref{fig:fig5}D, where ReLU nonlinearities are introduced. \\
\vspace{-.4cm}
\begin{tcolorbox}[colback=blue!3!white, colframe=blue!10!white, coltext=black, coltitle=black, title=\textbf{Takeaway: How can you maximise target alignment of PC?}
]
    Maximisation of target alignment in predictive coding can be achieved through:  
    \begin{itemize}
            \item \textbf{Initialisation.} Use initialisation that preserves the activity norm across layers.
            \item \textbf{Weight update rescaling.} Normalise the weight updates by: \\ - the dot product of pre and post inference activity for online learning, or \\ - the covariance of pre and post inference activity for batch training.
    \end{itemize}    
\end{tcolorbox}
\vspace{-.1cm}

\section{Conclusion and Future directions}
\vspace{-.1cm}
Inspired by early empirical reports of reduced destructive interference in PC relative to BP, we set out to provide a principled analytical account of when and why such advantages should arise. By deriving closed-form expressions for the changes in model predictions induced by BP and PC in DLNs, we show that the two learning rules differ fundamentally in how interference emerges during learning. In particular, BP updates are distorted by parameter-dependent interference arising from correlated and poorly conditioned weights, whereas PC attenuates interference, providing improved target alignment. We further show that, by introducing layer-dependent learning rates one can guarantee interference-free learning for single samples, independently of the number of inference steps, architecture or weight conditioning. Finally, extending the analysis to multi-sample learning reveals an additional source of cross-sample interference. In PC, this can be mitigated through activity-dependent decorrelation of weight updates, recovering interference-free learning trajectories.

\subsection{Limitations and future directions}

While our analysis provides useful theoretical and empirical insights, it has several limitations. First, the theory assumes linear networks, 
enabling tractability but not fully capturing the complexity of nonlinear models. Although preliminary experiments on nonlinear autoencoders suggest similar behavior, a rigorous extension to nonlinear settings remains open. Second, the analysis focuses on Multi-Layer Perceptrons. While the theory generalises to other linear models such as residual (Appendix \ref{sec:ch3_resnet}) and convolutional networks, we do not study these extensively. Extending target alignment analysis to broader architectures is left for future work. Finally, while weight or learning rate rescaling can accelerate PC, they introduce biologically implausible computations and are best viewed as optimal performance baselines. Biological systems may employ other strategies to maximise target alignment, such as activity normalisation \cite{carandini2012normalization}. Hence the presented theory can guide future investigation on how biological neural networks achieve their remarkable sample efficiency.

\newpage

\section*{Acknowledgements}
This work has been supported by the Medical Research Council UK grants MC\_UU\_00003/1, UKRI/MR/B000936/1, the Clarendon Fund, and by the Wellcome Trust grant 313955/Z/24/Z.

\section*{Code availability}
Code for all simulations described in the paper are freely available at: \url{https://github.com/EleneLomi/TAFigures.git}.


\bibliographystyle{unsrt}   
\bibliography{refs_new}

\medskip


\newpage
\appendix

\addcontentsline{toc}{section}{Appendix} 
\part{Appendix} 
\parttoc 


\section{Experimental Configurations}



\subsection{Training data}
\label{Appendix:TrainingData}

All experiments use synthetically generated data unless stated otherwise. Input vectors are sampled independently from a standard normal distribution, $x \sim \mathcal{N}(0, I)$, unless stated otherwise. Target outputs are generated via a fixed linear mapping,
\begin{equation*}
    y = W_{\text{data}} x,
\end{equation*}
where $W_{\text{data}}$ is sampled once per experiment seed with entries drawn independently from a zero-mean normal distribution with variance $1 / n_{\text{in}}$, where $n_{\text{in}}$ refers to the number of input neurons. This scaling ensures that output magnitudes remain comparable across different input dimensionalities. To train nonlinear models, we use the standard MNIST dataset.\\

\subsection{One training step experiments}

In this section we focus solely on the experiments carried out for a single training step. All experiments use fully connected DLNs. The default architecture consists of a linear model with 512 input neurons, a single hidden layer of 512 neurons, and 512 output neurons. This model is initialised using Kaiming uniform initialisation. The models are trained either on a single sample (Figure~\ref{fig:fig3}) or on batches of 128 samples (Figures~\ref{fig:fig2} and \ref{fig:fig4}), depending on the experiment.

We consider two initialisations in our experiments, Kaiming Uniform and Norm preservation (see section \ref{sec:Norm-preservingInit} for more details).
Under Kaiming uniform, each weight matrix $W_l \in \mathbb{R}^{m \times n}$ is sampled as $W_l \sim U\!\left(-\sqrt{\tfrac{1}{n}},\, \sqrt{\tfrac{1}{n}}\right)$. Under Norm-preserving normal initialisation, weights are sampled according to $W_l \sim \mathcal{N}\!\left(0,\, \tfrac{1}{m}\right)$.
Orthogonal initialisation is performed by setting the condition number of one as described below.

Architectural variations are introduced by independently varying network depth and width. Network depth is defined as the number of hidden layers and is varied over values ${1, \dots, 8}$. When varying depth, all hidden layers have equal width. Network width is varied over ${128, 256, 512, 1024}$ neurons, with input and output dimensions set to 512 neurons. 
The settings with 128 and 256 hidden neurons cause an information bottleneck. 
When width is varied, the model contains a single hidden layer.

To examine the effect of weight conditioning, we explicitly control the condition number of the weight matrices. Conditioning values are varied over ${1, 2, 10, 50, 10^3, 10^4, 10^5,}$ ${ 10^7, 10^9, 10^{12}}$. Conditioning is imposed by rescaling the singular values of initially Kaiming-initialised weight matrices while preserving their singular vectors, as described in section \ref{sec:conditioning_metric}.

Batch size is varied to study inter-sample interference effects. Batch sizes are varied across ${1, 32, 64, 128, 256, 480, 550, 1000, 2048}$. For experiments involving decorrelation matrices, batch sizes exceeding the smallest layer width are included to probe the limits of exact interference cancellation.

Across all experiments, parameters not explicitly varied are fixed to their default values. Results are averaged over multiple random seeds as specified in the corresponding figure captions.

\subsection{Training for multiple steps}
\label{sec:TrainingforMultipleSteps}

We conduct controlled experiments comparing backpropagation (BP) and predictive coding (PC), as well as their rescaled counterparts, on a synthetic supervised learning task where the target mapping is a known low-rank linear transformation as detailed in section \ref{Appendix:TrainingData}. For each run, we fix random seeds ranging from $0$ to $10$, pre-sample identical minibatches (set to either $1$ for online learning or $64$ for batch learning), and initialise all models with the same weights to ensure fair comparisons. We train multilayer perceptrons of varying depths and widths using stochastic gradient descent across a sweep of learning rates, and apply two different weight initialisation schemes, Kaiming Uniform and Norm Preservation. For each model, we choose the learning rate which results in the lowest error at the end of training to produce the plots of the training curves. The default architecture consists of a linear MLP with $20$ input neurons, $20$ output neurons, and a single hidden layer which we vary over $15, 20$ and $40$ hidden neurons. We also consider a network with $8$ hidden layers to study the effect of depth on training. Performance is assessed by computing the mean squared error between the model’s effective linear mapping - obtained by a product of its layer weights $W_{L:1}$ - and the ground-truth matrix $W_{\text{data}}$. This setup isolates differences in learning dynamics between BP and PC. All results are averaged over the $10$ random seeds before plotting.\\

A nonlinear autoencoder is also trained on MNIST reconstruction tasks using BP, PC, and their rescaled variants. The default architecture is a fully connected MLP with layer widths $(784, 128, 32, 128, 784)$. We use standard LeCun initialisation, and nonlinearities are imposed by ReLU activations after each hidden layer, while the output layer applies a sigmoid function. The architecture compresses the input into a 32-dimensional bottleneck representation before reconstructing it through a mirrored decoder. For a fast implementation on digital hardware, we use ePC \citep{goemaere2025erroroptimizationovercomingexponential}, a method developed to speed up the convergence of the inference phase in PC.\\

\section{Training with BP and PC}
\subsection{Training BP models}
\paragraph{Linear models.}
DLNs trained with BP are optimised by minimising a mean squared error objective using stochastic gradient descent. For a mini-batch of $B$ input-output pairs $\{(x_b, y_b)\}_{b=1}^B$, 
The continuous-time gradient descent dynamics for the weights of layer $l$ are given by
\begin{equation}
\frac{\partial W_l}{\partial t}
=
- \frac{1}{B}
\sum_{b=1}^{B}
\nabla_{W_l} \mathcal{L}_{\mathrm{BP}}
=
\frac{1}{B}
\sum_{b=1}^{B}
W_{L:l+1}^\top
\left(
y_b - W_{L:1} x_b
\right)
\left(
W_{l-1:1} x_b
\right)^\top ,
\end{equation}
where $W_{L:l+1}$ and $W_{l-1:1}$ denote the products of weight matrices above and below layer $l$, respectively, with the convention $W_{L:L+1} = W_{0:1} = I$.

In practice, these dynamics are discretised using an explicit Euler step, yielding the weight update
\begin{equation}
    \Delta W_l
    =
    \alpha
    \frac{1}{B}
    \sum_{b=1}^{B}
    W_{L:l+1}^\top
    \left(
    y_b - W_{L:1} x_b
    \right)
    \left(
    W_{l-1:1} x_b
    \right)^\top ,
\end{equation}
where $\alpha$ denotes the learning rate. We use a fixed learning rate of $\alpha = 10^{-4}$ in all experiments for a single training step.
This learning rule is commonly employed for training deep neural networks. Varying the learning rate does not affect the results.

For experiments illustrating learning trajectories in Figures \ref{fig:fig2}, \ref{fig:fig3}, and \ref{fig:fig4}, learning rate sweeps are performed over intervals shown in section \ref{sec:AdditionalExperimentalData}. Learning rates which achieve the lowest error by the end of training are picked for each model separately.


\paragraph{Non-linear models.}
Training in deep non-linear networks follows the same procedure, but with a non-linear forward mapping. The sample-wise loss function is given by
\begin{equation}
\mathcal{L}_{\mathrm{BP}}(x, y, W)
=
\frac{1}{2}
\left\lVert
y - f_W(x)
\right\rVert_2^2 ,
\end{equation}
where the network output is defined as
\begin{equation}
    f_W(x)
    =
    W_L f\!\left(
    W_{L-1} f\!\left(
    \cdots f\!\left(
    W_1 x
    \right)
    \right)
    \right),
\end{equation}
and $f(\cdot)$ denotes the ReLU activation function applied elementwise. Gradients are computed via standard BP through the non-linear activations, and the resulting weight updates are applied using the same discrete-time update rule and learning rate as in the linear case.

\subsection{Training PC models}
\paragraph{Linear models.}
DLNs trained with PC minimise local prediction errors by first iteratively updating the neural activity until convergence, followed by a single weight update with stochastic gradient descent.
In the experiments, the steady-state activity for each DLN layer $x_l^*$ is computed using the expression given in equation \ref{eq:ch3/pc_activivty_eq}. This is equivalent to iterative inference but is less computationally intensive.
For a mini-batch of $B$ input-output pairs, the continuous-time gradient descent dynamics for the weight of layer $l$ are given by
\begin{equation*}
    \frac{\partial W_l}{\partial t} = - \nabla_{W_l}{E} = \frac{1}{B}\sum_{b=1}^B W_{L:l+1}^\top\ S^{-1} (y_b-W_{L:1}x_b)\ {x}_{l-1,b}^{*\top}.
\end{equation*}

In the experiments, these dynamics are discretised using an explicit Euler step, yielding
\begin{equation}
    \Delta W_l = \alpha \frac{1}{B}\sum_{b=1}^B W_{L:l+1}^\top\ S^{-1} (y_b-W_{L:1}x_b)\ {x}_{l-1,b}^{*\top}.
    \label{eq:ch3_discPC_update}
\end{equation}
where $\alpha$ denotes the learning rate. We use a fixed learning rate of $\alpha = 10^{-4}$ in all experiments for one training step, unless stated otherwise. Varying the learning rate does not affect the results.

As for BP, for experiments illustrating learning trajectories with PC in Figures \ref{fig:fig2}, \ref{fig:fig3}, and \ref{fig:fig4}, learning rate sweeps are performed over intervals shown in section \ref{sec:AdditionalExperimentalData}. Learning rates which achieve the lowest error by the end of training are picked for each model separately.


\paragraph{Non-linear models.}
For deep non-linear networks, PC's energy function is given by 
\begin{equation*}
    E(x_0, ...,x_L, W) = \frac{1}{B}\sum_{b=1}^B \sum_{l=1}^L \frac{1}{2}||x_{l,b} - f(W_{l}x_{l-1,b})||^2_2.
\end{equation*}
where $f(\cdot)$ is a ReLU activation function. 
The feedforward pass of this model is identical to that of the corresponding non-linear model trained with BP. The training procedure otherwise follows that used for PC in DLNs. However, inference is no longer analytically tractable in the non-linear setting. Consequently, the converged neural activities $x_l^*$ are obtained via iterative inference. Inference is performed by minimising PC's energy function using gradient descent for $10,000$ steps, with an inference step of $0.05$. We use a large number of inference steps to ensure that the neural activity reaches steady-state.



\section{Mechanisms for improving alignment}
\label{sec:alignment_factors}

In order to maintain high target alignment, the norm of layer activities should remain approximately constant throughout training as mentioned in section \ref{sec:HigherTAinPC}. A loose constraint to encourage this is initialising networks with a Norm-preserving initialisation. We discuss this initialisation in section \ref{sec:Norm-preservingInit} and show why the default Kaiming Uniform initialisation for linear networks violates activity norm preservation. In several experiments, we introduce tighter constraints where we modify the standard PC and BP weight updates via additional multiplicative factors. These factors are designed to reduce interference, thereby improve target alignment. Two classes of factors are considered: scalar, activity-dependent learning factors for single-sample training, and matrix-valued decorrelation factors for batch training. In sections \ref{sec:ActivityDependentLearningRates} and \ref{sec:DecorrelationMatrixFactors}, we give a detailed account of these.

\subsection{Different initialisation schemes}
\label{sec:Norm-preservingInit}

In this section, we show the requirements to preserve the norm of the layers at initialisation and show why the default Kaiming initialisation in pytorch violates this.

\subsubsection{Norm-preserving initialisation}

The goal is to preserve the expected squared norm
\begin{equation}
\mathbb{E}\|Wx\|^2 = \mathbb{E}\|x\|^2,
\end{equation}

where $x \in \mathbb{R}^n$, with $x_i \sim \mathcal{N}(0,1)$ and $W \in \mathbb{R}^{m \times n}$, with $W_{ij} \sim \mathcal{N}(0, \sigma^2)$.

Each output component is $(Wx)_i = \sum_{j=1}^{n} W_{ij} x_j.$ Since everything is zero-mean and independent, we obtain

\begin{equation}
\mathrm{Var}((Wx)_i) = \sum_{j=1}^{n} \mathrm{Var}(W_{ij} x_j) = \sum_{j=1}^{n} \sigma^2 = n\sigma^2.
\end{equation}

This leads us to the following expressions:
\begin{equation}
\mathbb{E}\|Wx\|^2 = \sum_{i=1}^{m} \mathbb{E}[(Wx)_i^2] = m \cdot n \sigma^2,
\end{equation}

\begin{equation}
\mathbb{E}\|x\|^2 = n.
\end{equation}

Equating the expectations of both expressions,  we obtain the final result:

\begin{equation}
\sigma = \frac{1}{\sqrt{m}}.
\end{equation}

For a square network, since $m=n$ the variance of individual neurons is also preserved since, letting $z_j=\sum_{i=1}^n W_{ji}x_i$, we have that Var$(z_j)=\mathrm{Var}(x_i)$ as long as $\sigma = \frac{1}{\sqrt{n}}$. Norm-preserving initialisation does not generally preserve variance at neuron level for arbitrary architectures, which may lead to instabilities during training for extreme bottlenecks or very wide hidden layers.\\

\subsubsection{Kaiming Uniform initialisation and activity-norm decay}

For default Kaiming initialisation, we again have $x \in \mathbb{R}^n$, with $x_i \sim \mathcal{N}(0,1)$ and $W \in \mathbb{R}^{m \times n}$, with $W_{ij} \sim \mathcal{U}(-\sqrt{\frac{1}{n}}, \sqrt{\frac{1}{n}}).$

Then
\begin{equation}
\mathbb{E}[W_{ij}] = 0, 
\qquad 
\mathrm{Var}(W_{ij}) = \mathbb{E}[W_{ij}^2] = 
\frac{1}{3n}.
\end{equation}

This leads us to the following:
\begin{align}
\mathbb{E}\|Wx\|^2 &= \mathbb{E}[x^\top W^\top W x] = \mathbb{E}\big[x^\top \mathbb{E}[W^\top W] x\big] \\
&= \frac{m}{3n} \, \mathbb{E}\|x\|^2
\end{align}

because $\mathbb{E}[W^\top W] 
= \sum_{i=1}^m \mathbb{E}[w_i w_i^\top] 
= m \, \mathbb{E}[W_{ij}^2] \, I 
= m \frac{1}{3n} I$. $w_i$ denotes the $i$-th row of $W$.

Therefore, overall we have that
\begin{equation}
\mathbb{E}\|Wx\|^2 = \frac{m}{3n} \, \mathbb{E}\|x\|^2.
\end{equation}
Considering square networks where $m=n$, we can see that the layer norm of the activities decays in expectation throughout the forward pass by a factor of $3$.

\subsection{Activity-dependent learning rates}
\label{sec:ActivityDependentLearningRates}
We first introduce layer-specific scalar factors that rescale the weight updates at each layer. For PC, the continuous-time weight update at layer $l$ is modified as
\begin{equation}
\frac{\partial W_l}{\partial t} = - \alpha_l \nabla_{W_l} E.
\end{equation}

The scaling factor $\alpha_l$ is chosen to cancel the activity-dependent scalar term appearing in PC’s change in prediction and is defined as
\begin{equation}
\alpha_l = \frac{1}{x^{*\top}_{l-1} \hat{x}_{l-1}},
\end{equation}
where $\hat{x}_{l-1}$ denotes the feedforward activity at layer $l-1$ and $x^*_{l-1}$ denotes the converged activity following PC inference. 

For comparison, we also apply analogous activity-dependent scaling to BP. In this case, the scaling factor at layer $l$ is given by
\begin{equation}
\alpha_l = \frac{1}{\hat{x}^\top_{l-1} \hat{x}_{l-1}},
\end{equation}
which removes the corresponding activity-dependent scalar factor from BP’s prediction update.

Introducing these activity-dependent learning factors is equivalent to using
layer-dependent learning rates. For instance, for networks trained with PC, the
discretised weight updates given in equation~\ref{eq:ch3_discPC_update} become
\begin{equation*}
    \Delta W_l
    = \alpha_l \alpha 
    W_{L:l+1}^\top S^{-1}
    (y - W_{L:1} x)\,
    x^{*\top}_{l-1}.
\end{equation*}
Here, $\alpha$ denotes the global learning rate and $\alpha_l$ is a layer-specific
scaling factor. Their product can therefore be interpreted as an effective learning
rate for layer $l$.

In all experiments employing these scaling factors, the layer-specific coefficients $\alpha_l$ are recomputed for each sample and applied only for the duration of a single weight update. Apart from the inclusion of these factors, the training procedure is identical to that used without scaling. In particular, all experiments use the same global learning rate and the same number of inference steps.

\subsection{Decorrelation matrix factors}
\label{sec:DecorrelationMatrixFactors}
We next introduce layer-specific {matrix-valued} factors that rescale the weight
updates to remove inter-sample interference. These factors are designed to cancel the activity-dependent matrix
terms that appear in the change in prediction when learning from multiple samples,
thereby improving target alignment.

For PC, the continuous-time weight update at layer $l$ is equal to
\begin{equation}
    \frac{\partial W_l}{\partial t} =-\frac{1}{B}\sum_{b=1}^{B}\nabla_{W_l} {E}\, A_l ,
\end{equation}
where $A_l$ is a decorrelation matrix applied to the right of the gradient.

The decorrelation matrix is chosen to cancel the batch-averaged activity correlation
term appearing in PC’s change in prediction. Specifically, we define
\begin{equation}
    A_l=\Big( E_B\big[\hat x_{l-1,b}\,{x}_{l-1,b}^{*\top}\big]\Big)^{+},
\end{equation}
where $\hat{x}_{l-1,b}$ denotes the feedforward activity at layer $l-1$ for sample $b$,
$x^*_{l-1,b}$ denotes the corresponding converged activity following PC inference.
$E_B[\cdot]$ denotes the expectation over batch elements calculated with the empirical mean across the batch, and $.^+$ denotes the pseudo inverse.

For comparison, we also apply an analogous decorrelation factor to BP.
In this case, the matrix factor is defined using feedforward activities only:
\begin{equation}
    A_l=\Big(E_B\big[\hat{x}_{l-1,b}\,\hat{x}_{l-1,b}^\top\big]\Big)^{+}.
\end{equation}

These matrix-valued factors are recomputed separately for each layer at every weight update. Matrix inversion is performed using the Moore–Penrose pseudoinverse to maintain numerical stability. The training procedure remains unchanged with identical training parameters to those used in the standard settings.

\section{Metrics}
\subsection{Target alignment}
\label{sec:target_alignment_metric}
Target alignment quantifies how closely a learning update follows the direction of the output prediction error. Target alignment is defined as the cosine similarity between the output residual $r = y - \hat{y}$ and the change in the model’s prediction induced by a weight update, $\Delta \hat{y}$ (see equation \ref{eq:ch3_target_alignmet_eq}).
A value of one indicates perfect alignment with the residual, zero indicates orthogonality, and negative values indicate updates that move the prediction away from the target. In experiments involving mini-batches, target alignment is computed separately for each sample and reported as the mean across samples. All reported results correspond to the alignment induced by a single parameter update.

\subsection{Condition number}
\label{sec:conditioning_metric}

The conditioning of a model’s weights is quantified using the condition number of each weight matrix. For a weight matrix $W_l$, the condition number is defined as
\begin{equation}
\kappa(W_l) = \frac{\sigma_{\max}(W_l)}{\sigma_{\min}(W_l)},
\end{equation}
where $\sigma_{\max}$ and $\sigma_{\min}$ denote the largest and smallest singular values, respectively. A condition number of one corresponds to perfectly conditioned (orthogonal) weights, while larger values indicate poorer conditioning and increased susceptibility to interference. When conditioning is varied experimentally, all layers are assigned the same condition number. 

We set the conditioning by modifying the singular value spectrum of each weight matrix while preserving its singular vectors. For a given layer $l$, we compute the singular value decomposition
\begin{equation}
W_l = U_l \Sigma_l W_l^\top,
\end{equation}
where $\Sigma_l = \mathrm{diag}(\sigma_1,\dots,\sigma_k)$ contains the singular values in descending order. We then replace the original singular values by a linearly spaced spectrum
\begin{equation}
\sigma_i' = \sigma_{\max}\left(1 - \frac{i-1}{k-1}\left(1 - \frac{1}{\kappa}\right)\right),
\end{equation}
where $\sigma_{\max}$ is the largest original singular value, $k$ is the rank of $W_l$, and $\kappa$ is the desired condition number. This construction ensures $\sigma_{\max}' / \sigma_{\min}' = \kappa$.

To isolate the effect of conditioning from changes in the overall weight scale, the modified singular values, $\Sigma'_l$, are rescaled to preserve the Frobenius norm of the original weight matrix,
\begin{equation} 
\Sigma_l' \leftarrow \Sigma_l' \frac{\|\Sigma_l\|_F}{\|\Sigma_l'\|_F}.
\end{equation}
The conditioned weight matrix is then reconstructed as $W_l \leftarrow U_l \Sigma_l' W_l^\top$. This procedure alters only the conditioning of the weights while keeping their total energy and singular directions fixed.

\section{Derivation of PC's weight updates\label{s:ch3_deriV_pc_update}}

PC updates its weights by following the negative gradient of its energy function with respect to the synaptic parameters. For the PC energy function
\begin{equation*}
    E(x_0,...,x_L, W) = \sum_{l=1}^L \frac{1}{2}\|x_l - W_l x_{l-1}\|_2^2,
\end{equation*}
the weight dynamics are given by
\begin{equation*}
    \frac{\partial W_l}{\partial t}
    = - \nabla_{W_l} E
    = (x_l^* - W_l x_{l-1}^*) x_{l-1}^{*\top}
    = \epsilon_l^* x_{l-1}^{*\top},
\end{equation*}
where $x_l^*$ denotes the equilibrium activity of layer $l$, and $\epsilon_l^*$ is the corresponding prediction error at equilibrium. The full expression for $x^*_l$ is given below and derived in \cite{ishikawa2025locallossoptimizationinfinite}:


\begin{equation}
    x_l^* = \underbrace{W_{l:1} x}_{\text{BP feed forward activity}} + \underbrace{\Big(W_{L :l+1}^\top + \sum_{i=2}^{l} W_{l:i}W_{L:i}^\top\Big)  \Big(I + \sum_{i=2}^{L} W_{L:i}W_{L:i}^\top\Big)^{-1} r}_{\text{propagated errors}},  
    \label{eq:ch3/pc_activivty_eq}
\end{equation}

$x_l^*$ can be seen as a sum of BP’s feedforward pass activity values and the errors propagated backwards 
\cite{millidge2022theoreticalframeworkinferencelearning, ishikawa2025locallossoptimizationinfinite}.
At the activity equilibrium reached during inference, the gradient of the energy with respect to the neural activities vanishes. This yields, for $l = 2, \dots, L-1$,
\begin{equation*}
    \nabla_{x_l} E
    = (x_l^* - W_l x_{l-1}^*) - W_{l+1}^\top (x_{l+1}^* - W_{l+1} x_l^*)
    = 0.
\end{equation*}
Rearranging this expression gives
\begin{align*}
    (x_l^* - W_l x_{l-1}^*) &= W_{l+1}^\top (x_{l+1}^* - W_{l+1} x_l^*), \\
    \epsilon_l^* &= W_{l+1}^\top \epsilon_{l+1}^*.
\end{align*}
By recursively applying this relation, the prediction error at layer $l$ can be expressed in terms of the output-layer error. Substituting into the weight update yields
\begin{equation}
    \frac{\partial W_l}{\partial t}
    = \epsilon_l^* x_{l-1}^{*\top}
    = W_{L:l+1}^\top \epsilon_L^* x_{l-1}^{*\top},
    \label{eq:ch3_methods_other_one}
\end{equation}
where $W_{L:l+1} = W_L W_{L-1} \cdots W_{l+1}$.

The equilibrium activity can be written equivalently as
\begin{equation*}
    x_l^* = W_l x_{l-1}^* + \epsilon_l^*.
\end{equation*}
Using this recursive form, the output activity can be expanded as
\begin{align*}
    x_L^*
    &= W_L x_{L-1}^* + \epsilon_L^* \\
    &= W_L (W_{L-1} x_{L-2}^* + \epsilon_{L-1}^*) + \epsilon_L^* \\
    &= W_{L:1} x_0^* + W_{L:2} \epsilon_1^* + W_{L:3} \epsilon_2^* + \dots + \epsilon_L^* \\
    &= W_{L:1} x_0^* + \sum_{l=2}^L W_{L:l} \epsilon_{l-1}^* + \epsilon_L^* .
\end{align*}

Since the input and output layers are clamped to the input $x$ and target $y$, respectively, and using the recursive expression for the prediction errors, the above expression becomes
\begin{align*}
    x_L^*
    &= W_{L:1} x_0^* + \sum_{l=2}^L W_{L:l} \epsilon_{l-1}^* + \epsilon_L^*, \\
    y
    &= W_{L:1} x + \sum_{l=2}^L W_{L:l} W_{L:l}^\top \epsilon_L^* + \epsilon_L^*, \\
    y - W_{L:1} x
    &= \Bigl(I + \sum_{l=2}^L W_{L:l} W_{L:l}^\top \Bigr) \epsilon_L^* .
\end{align*}

Using the definitions of $S$ and $r$ introduced in the main text, this simplifies to
\begin{align}
    r &= S \, \epsilon_L^*, \\
    \epsilon_L^* &= S^{-1} r .
    \label{eq:ch3_methods_el_S}
\end{align}

Finally, combining equations~\ref{eq:ch3_methods_other_one} and~\ref{eq:ch3_methods_el_S}, we obtain the expression used in the manuscript for the PC weight updates (equation \ref{ch3/weightupdates}):
\begin{equation*}
\boxed{
    \frac{\partial W_l}{\partial t}
    = W_{L:l+1}^\top \epsilon_L^* x_{l-1}^{*\top}
    = W_{L:l+1}^\top S^{-1} r \, x_{l-1}^{*\top}.
}
\end{equation*}

\section{Proof of Theorem \ref{theorem1}: target alignment for adaptive learning rates} \label{appendix:thorem1}

Let us remind that theorem \ref{theorem1} assumes that we modify PC's weight update to include a layer-specific scaling factor $\alpha_l$:
\begin{equation}
    \frac{\partial W_l}{\partial t} = -\alpha_l \nabla_{W_l}{E}
\end{equation}
In practice, this is equivalent to assigning different learning rates for each layer.
Choosing $\alpha_l$ equal to the inverse of PC's scalar factor $x_{l-1}^{*\top}\hat x_{l-1}$ removes the dependency of PC's change in prediction on this scalar factor. This allows the expression for PC's change in prediction to be aligned with the residual $r$, leading to maximal target alignment as shown below: 
\begin{align}
    \frac{\partial \hat y}{\partial t}
    &=   \sum_{l=1}^L W_{L:l+1}\frac{\partial W_l}{\partial t} W_{l-1:1}x =   -\sum_{l=1}^L W_{L:l+1}\ \alpha_l \nabla_{W_l}{E}\ \hat x_{l-1}\\
    &=   \sum_{l=1}^L W_{L:l+1} (1/{x}_{l-1}^{*\top} \hat x_{l-1}) W_{L:l+1}^\top\ S^{-1} r\ {x}_{l-1}^{*\top} \hat x_{l-1} \\
    &=   \underbrace{\sum_{l=1}^L W_{L:l+1} W_{L:l+1}^\top\ S^{-1}}_{=I}\ \quad  r \\&= r
    \label{eq:ch3_PC_scaling}
\end{align}

\section{Changes in prediction for batch training} \label{appendix_batch_train}
We re-derive the change in predictions of a DLN when training on a mini-batch of data. For batch training, weight updates are the average of the per-sample updates across the batch.

\paragraph{Backpropagation.}
For BP with a mini-batch of size $B$, the weight update for layer $l$ is
\begin{equation}
    \frac{\partial W_l}{\partial t}
    = -\frac{1}{B}\sum_{b=1}^B \nabla_{W_l}\mathcal{L}_{\mathrm{BP}}
    = \frac{1}{B}\sum_{b=1}^B W_{L:l+1}^\top r_b \, \hat{x}_{l-1,b}^\top
    = \frac{1}{B} W_{L:l+1}^\top R \, \hat{X}_{l-1}^\top .
\end{equation}
where $x_b$ and $y_b$ denote the input and target output for the $b$-th data point in the batch, and $r_b$ is the corresponding output prediction residual. For compactness, we define $R = [r_1,\dots,r_B]$ as the matrix of residuals and $\hat{X}_{l-1} = [\hat{x}_{l-1,1},\dots,\hat{x}_{l-1,B}]$ as the matrix of feedforward activities at layer $l-1$.

The resulting change in the predictions across the batch is
\begin{equation}
    \frac{\partial \hat{Y}}{\partial t}
    = \frac{1}{B}\sum_{l=1}^L
    W_{L:l+1} W_{L:l+1}^\top
    R \ \hat{X}_{l-1}^\top \hat{X}_{l-1},
\end{equation}
where $ \hat{Y} = [ \hat{y}_1,\dots, \hat{y}_B]$ collects the predictions for all batch elements.

\paragraph{Predictive coding.}
For PC trained on a mini-batch, the weight update for layer $l$ is given by
\begin{equation}
    \frac{\partial W_l}{\partial t}
    = -\frac{1}{B}\sum_{b=1}^B \nabla_{W_l} {E}
    = W_{L:l+1}^\top S^{-1} R \, X_{l-1}^{*\top}.
\end{equation}
with $X_{l-1}^{*} = [x_{l-1,1}^{*},\dots,x_{l-1,B}^{*}]$.

The corresponding change in PC's predictions is
\begin{equation}
    \frac{\partial \hat{Y}}{\partial t}
    = \frac{1}{B}\sum_{l=1}^L
    W_{L:l+1} W_{L:l+1}^\top
    S^{-1} R \ X_{l-1}^{*\top} \hat{X}_{l-1}.
\end{equation}

\section{Proof of Theorem \ref{theorem2}: mitigating batch interference in PC}
\label{appendix_theorem2}






Let us first remind the theorem.
Define the layer-wise rescaling factor
\begin{equation}
  A_l \;=\; \Big(\tfrac1B\textstyle\sum_{b=1}^{B}\hat x_{l-1,b}\,x^{*\top}_{l-1,b}\Big)^{+}
        \;=\; \big(\tfrac1B\,\hat X_{l-1}X^{*\top}_{l-1}\big)^{+}.
\end{equation}
If the batch size satisfies $B\le d_{l}$ for every rescaled layer $l=0,\dots,L-1$ and the batch activities are in general position (i.e., $\hat X_{l-1}$ and $X^{*}_{l-1}$ have full column rank $B$), then the weight update
$\partial W_l/\partial t = -\tfrac1B\sum_b \nabla_{W_l}E\;A_l$ yields, for every
sample in the batch, $\partial \hat y_b/\partial t \propto r_b$, i.e.\ optimal
target alignment holds \emph{exactly and simultaneously for all $b$}.

\paragraph{Proof.} 
The rescaled change in prediction is 
\begin{equation}
\frac{\partial \hat Y}{\partial t}=\frac1B\sum_{l}W_{L:l+1}W_{L:l+1}^\top S^{-1}R\, \left[X^{*\top}_{l-1}A_l\hat X_{l-1}\right].
\label{eq:dY}
\end{equation}
Exact per-sample alignment requires the
inter-sample interference matrix to be a (scalar multiple of the) identity,
i.e. the term in square brackets in the equation above needs to satisfy $X^{*\top}_{l-1}A_l\hat X_{l-1}=c\,I_B$, where $I_B$ denotes an identity matrix of size $B$. Under $B\le min_{l=0,\dots,L-1} d_{l}$ and general position,
$\hat X_{l-1}$ has full column rank and $X^{*\top}_{l-1}$ has full row rank, so the
reverse-order pseudo-inverse law applies exact  ly,
$(\hat X_{l-1}X^{*\top}_{l-1})^{+}=(X^{*\top}_{l-1})^{+}(\hat X_{l-1})^{+}$, and
the one-sided identities $X^{*\top}_{l-1}(X^{*\top}_{l-1})^{+}=I_B$,
$(\hat X_{l-1})^{+}\hat X_{l-1}=I_B$ give
\begin{equation}
  X^{*\top}_{l-1}A_l\hat X_{l-1}  = B \big[ X^{*\top}_{l-1}(X^{*\top}_{l-1})^{+} \big] \big[ (\hat X_{l-1})^{+} \hat{X}_{l-1} \big]= B I_B. 
  \label{eq:BIB}
\end{equation}
Substituting equation \ref{eq:BIB} into equation \ref{eq:dY} and using $\sum_l W_{L:l+1}W_{L:l+1}^\top=S$, we obtain:
\begin{equation}
\frac{\partial\hat Y}{\partial t}=\frac1B\,[S S^{-1}]\,R\,(B) = R.
\end{equation}
This implies that for every training sample in the batch:
$\partial\hat y_b/\partial t=r_b$. The global scalar $c=B$ cancels the $1/B$; in
general it is absorbed into the learning rate, and target alignment (a cosine) is
scale-invariant. \hfill$\square$

\paragraph{Necessity and the $B>\text{width}$ regime.} The condition
$B\le \min_l d_{l-1}$ is also necessary: $X^{*\top}_{l-1}A_l\hat X_{l-1}$ factors
through $\mathbb R^{d_{l-1}}$, so $\operatorname{rank}(X^{*\top}_{l-1}A_l\hat
X_{l-1})\le d_{l-1}$ for \emph{any} $A_l$. If $B>d_{l-1}$ then $\operatorname{rank}(cI_B)=B>d_{l-1}$ is unattainable. This is the departure from alignment past the dashed line (network width) in Figure~\ref{fig:fig4}A.

Table \ref{tab:pinv} experimentally validates the use of pseudo-inverses and their conditions on the batch size.
\begin{table}[h]
\centering
\caption{Deviation of $X^{*\top}A\hat X$ from a scalar multiple of the identity
($0\Rightarrow$ optimal per-sample target alignment). 
}
\label{tab:pinv}
\begin{tabular}{ccl|c|c}
\hline
$d$ & $B$ & regime & code order $(\hat X X^{*\top})^{+}$ & reversed order $(X^{*}\hat X^{\top})^{+}$\\
\hline
64  & 16 & $B\le d$          & $3.2\times10^{-15}$ & $5.0\times10^{-1}$\\
128 & 32 & $B\le d$          & $2.5\times10^{-15}$ & $4.6\times10^{-1}$\\
64  & 63 & $B\le d$          & $8.3\times10^{-14}$ & $1.9\times10^{1}$\\
64  & 64 & $B=d$             & $7.0\times10^{-14}$ & $3.8\times10^{2}$\\
64  & 80 & $B>d$ (rank-lim.) & $2.4$               & $3.9\times10^{1}$\\
16  & 64 & $B>d$ (rank-lim.) & $2.1$               & $2.4$\\
\hline
\end{tabular}
\end{table}

\section{Proof of Theorem \ref{theorem:finite_T_alignment}: target alignment without equilibrated energy}
\label{proof:3}
\begin{proof}
We first prove the theorem for batch size $B$, followed by a special case of online learning $B=1$.

\paragraph{Batch learning.}
Let $\hat{x}_{0,b}=x_b,$ and let $\hat{x}_{l,b}=W_l\hat{x}_{l-1,b}$ be the feedforward activities. Then the prediction is $\hat{y}_b=\hat{x}_{L,b}.$\\

After $T$ inference steps, the prediction errors are
\begin{equation}
    \epsilon_{l,b}^{(T)}
    =
    x_{l,b}^{(T)}
    -
    W_lx_{l-1,b}^{(T)}.
\end{equation}
Since the input and target activities are clamped,
$x_{0,b}^{(T)}=x_b$ and $x_{L,b}^{(T)}=y_b$.

Collect the feedforward and inferred activities
and prediction errors into matrices
\begin{align}
    \hat{X}_{l-1}
    &=
    [\hat{x}_{l-1,1},\ldots,\hat{x}_{l-1,B}],
    \\
    X_{l-1}^{(T)}
    &=
    [x_{l-1,1}^{(T)},\ldots,x_{l-1,B}^{(T)}],
    \\
    \mathcal{E}_l^{(T)}
    &=
    [\epsilon_{l,1}^{(T)},\ldots,
      \epsilon_{l,B}^{(T)}].
\end{align}

The partial gradient with respect to $W_l$ is
\begin{equation}
    \nabla_{W_l}E_b^{(T)}
    =
    -\epsilon_{l,b}^{(T)}
    x_{l-1,b}^{(T)\top}.
\end{equation}

In the matrix notation introduced above, the rescaled weight update assumed in Theorem \ref{theorem:finite_T_alignment} can be written as:
\begin{equation}
    \frac{\partial W_l}{\partial t}
    =
    \frac{1}{B}
    \mathcal{E}_l^{(T)}
    X_{l-1}^{(T)\top}A_l^{(T)}.
    \label{eq:rescaled_update_T}
\end{equation}

To simplify expressions, we write $W_LW_{L-1}\cdots W_{l+1}=W_{L:l+1}$ (note $W_{L:L+1}$ is set to identity).

For a linear network, differentiating the
feedforward prediction with respect to learning
time gives
\begin{equation}
    \frac{\partial\hat{Y}}{\partial t}
    =
    \sum_{l=1}^{L}
    W_{L:l+1}
    \frac{\partial W_l}{\partial t}
    \hat{X}_{l-1},
    \label{eq:prediction_dynamics_T}
\end{equation}
where
$\hat{Y}=[\hat{y}_1,\ldots,\hat{y}_B]$.

Substituting Eq.~\eqref{eq:rescaled_update_T}
into Eq.~\eqref{eq:prediction_dynamics_T},
we obtain
\begin{align}
    \frac{\partial\hat{Y}}{\partial t}
    &=
    \sum_{l=1}^{L}
    W_{L:l+1}\mathcal{E}_l^{(T)}
    \left[
    \frac{1}{B}
    X_{l-1}^{(T)\top}
    A_l^{(T)}
    \hat{X}_{l-1}
    \right].
    \label{eq:batch_prediction_T}
\end{align}

We can make the expression in square brackets
equal to the identity matrix $I_B$ by choosing
\begin{equation}
    A_l^{(T)}
    =
    B
    \left(X_{l-1}^{(T)\top}\right)^+
    \hat{X}_{l-1}^+,
    \label{eq:correct_batch_A_T}
\end{equation}
where $(\cdot)^+$ denotes the Moore--Penrose
pseudoinverse.

By assumption, both $X_{l-1}^{(T)}$ and
$\hat{X}_{l-1}$ have full column rank $B$ (we assume the batch size is less than or equal to the minimum layer width in the network $B\leq\textrm{min}_ld_l$).
Therefore,
\begin{equation}
    X_{l-1}^{(T)\top}
    \left(X_{l-1}^{(T)\top}\right)^+
    =I_B
\end{equation}
and
\begin{equation}
    \hat{X}_{l-1}^+\hat{X}_{l-1}
    =I_B.
\end{equation}

Also note that the prediction errors satisfy the telescoping identity 
\begin{align}
    \sum_{l=1}^{L}W_{L:l+1}\epsilon_{l,b}^{(T)}
    &=
    \sum_{l=1}^{L}
    W_{L:l+1}\left(
    x_{l,b}^{(T)}
    -W_lx_{l-1,b}^{(T)}
    \right),
    \nonumber
\end{align}
expanding the brackets, we observe that all terms except initial and final cancel each other out, and we obtain: 
\begin{align}
    \sum_{l=1}^{L}W_{L:l+1}\epsilon_{l,b}^{(T)}
    &=y_b-W_L\cdots W_1x_b
    =y_b-\hat{y}_b
    =r_b.
    \label{eq:telescoping_T}
\end{align}
This identity holds independently of the number
of inference steps $T$.

Substituting Eq.~\eqref{eq:correct_batch_A_T},
we find
\begin{align}
    &\frac{1}{B}
    X_{l-1}^{(T)\top}
    A_l^{(T)}
    \hat{X}_{l-1}
    =
    X_{l-1}^{(T)\top}
    \left(X_{l-1}^{(T)\top}\right)^+
    \hat{X}_{l-1}^+
    \hat{X}_{l-1}
    =I_B I_B
    =I_B.
    \label{eq:batch_identity_T}
\end{align}

Using Eq.~\eqref{eq:batch_identity_T}
in Eq.~\eqref{eq:batch_prediction_T}
gives
\begin{align}
    \frac{\partial\hat{Y}}{\partial t}
    =
    \sum_{l=1}^{L}
    W_{L:l+1}\mathcal{E}_l^{(T)}
    =Y-\hat{Y},
\end{align}
where the last equality follows from
the telescoping identity
in Eq.~\eqref{eq:telescoping_T},
applied to all $B$ samples.

Thus, for every sample $b=1,\ldots,B$,
\begin{equation}
    \boxed{
    \frac{\partial\hat{y}_b}{\partial t}
    =y_b-\hat{y}_b=r_b.
    }
\end{equation}

Hence, the instantaneous
change in the feedforward prediction is
perfectly aligned with its target residual.
Since the telescoping identity holds for
all $T\geq 0$, the result does not require
inference to converge to equilibrium.

\paragraph{Online learning.}
For online learning, $B=1$. In this case,
consider the scalar rescaling
\begin{equation}
    \alpha_l^{(T)}
    =
    \frac{1}{
    x_{l-1}^{(T)\top}\hat{x}_{l-1}
    },
    \label{eq:online_scaling_T}
\end{equation}
assuming
$x_{l-1}^{(T)\top}\hat{x}_{l-1}\neq 0$.

The rescaled local weight update becomes
\begin{equation}
    \frac{\partial W_l}{\partial t}
    =
    \alpha_l^{(T)}
    \epsilon_l^{(T)}
    x_{l-1}^{(T)\top}.
\end{equation}

The induced change in the feedforward prediction is
\begin{align}
    \frac{\partial\hat{y}}{\partial t}
    &=
    \sum_{l=1}^{L}
    W_{L:l+1}\frac{\partial W_l}{\partial t}
    \hat{x}_{l-1}
    \nonumber\\
    &=
    \sum_{l=1}^{L}
    W_{L:l+1}\epsilon_l^{(T)}
    \frac{
    x_{l-1}^{(T)\top}\hat{x}_{l-1}
    }{
    x_{l-1}^{(T)\top}\hat{x}_{l-1}
    }
    \nonumber\\
    &=
    \sum_{l=1}^{L}
    W_{L:l+1}\epsilon_l^{(T)}
    \nonumber\\
    &=y-\hat{y}=r.
\end{align}

Therefore, online learning also satisfies
\begin{equation}
    \boxed{
    \frac{\partial\hat{y}}{\partial t}=r
    }
\end{equation}
for any finite number of inference steps $T$,
provided the rescaling denominators are nonzero.\\

Therefore, under our assumptions rescaled predictive
coding achieves perfect instantaneous target
alignment independently of inference convergence for batch and online learning.
\end{proof}

\section{Extension to linear residual networks }
\label{sec:ch3_resnet}

\newcommand{\tildeb}[1]{\stackrel{\sim}{\smash{#1}\rule{0pt}{1.1ex}}}

The analytical framework developed for DLNs extends directly to linear residual networks (ResNets). We briefly summarize these results, which demonstrate that the fundamental mechanisms of PC interference reduction remain unchanged under skip connections.

In linear ResNets, each layer composes as $x_l = x_{l-1} + W_l x_{l-1} = (I + W_l)x_{l-1}$, where weights accumulate as $\tildeb{W} = \prod_{k=1}^l (I+W_k)$ rather than as pure products $W_{l-1:1}$.

\begin{figure}[h]
    \centering
    \includegraphics[width=0.6\linewidth]{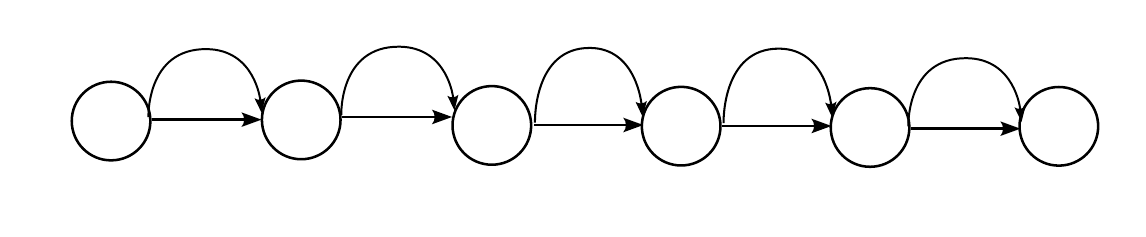}
    \caption*{}
    \label{fig:resnet}
\end{figure}

For clarity, we derive the ResNet expressions in the same order used above for DLNs: first the weight dynamics, then the induced change in prediction.

\textbf{Backpropagation}

For linear ResNets with $\hat y = \tildeb{W}_{L-1:1}x$ and residual $r = y-\hat y$, the continuous-time BP weight dynamics are
\begin{equation}
    \frac{\partial W_l}{\partial t}
    = \tildeb{W}_{L-1:l+1}^\top r\, \hat x_{l-1}^\top,
\end{equation}
where $\tildeb{W}_{L-1:l+1} = \prod_{k=l+1}^{L-1}(I+W_k)$ and $\hat x_{l-1}=\tildeb{W}_{l-1:1}x$.

The change in prediction then follows from product-rule differentiation of $\tildeb{W}_{L-1:1}$:
\begin{align}
    \frac{\partial \hat y}{\partial t}
    &= \frac{\partial \tildeb{W}_{L-1:1}}{\partial t}x \\
    &= \sum_{l=1}^{L-1} \tildeb{W}_{L-1:l+1}\frac{\partial W_l}{\partial t}\tildeb{W}_{l-1:1}x \\
    &= \sum_{l=1}^{L-1} \tildeb{W}_{L-1:l+1}\tildeb{W}_{L-1:l+1}^\top r\, \hat x_{l-1}^\top\hat x_{l-1}.
\end{align}
Hence, as in DLNs, interference is introduced by the matrix factor $\tildeb{W}_{L-1:l+1}\tildeb{W}_{L-1:l+1}^\top$ acting on $r$.

\textbf{Predictive coding}

For PC in ResNets, we start from the ResNet prediction-error objective
\begin{equation}
    E^{ResNet}(x_1,...,x_L,W)=\sum_{l=1}^{L-1}\frac{1}{2}\left\|x_{l+1}-(I+W_l)x_l\right\|_2^2.
\end{equation}
Define layer-wise prediction errors
\begin{equation}
    \epsilon_l := x_{l+1}-(I+W_l)x_l, \qquad l=1,\dots,L-1,
\end{equation}
so that $E^{ResNet}=\frac{1}{2}\sum_{l=1}^{L-1}\|\epsilon_l\|_2^2$. At inference equilibrium (with $x_1$ and $x_L=y$ clamped), stationarity with respect to hidden activities gives the recursion
\begin{equation}
    \epsilon_l = (I+W_{l+1})^\top \epsilon_{l+1}, \qquad l=1,\dots,L-2,
\end{equation}
which implies
\begin{equation}
    \epsilon_l = \tildeb{W}_{L-1:l+1}^\top\epsilon_{L-1}.
\end{equation}
The output residual can be written as
\begin{equation}
    r := y-\tildeb{W}_{L-1:1}x_1 = \epsilon_{L-1}+\sum_{l=1}^{L-2}\tildeb{W}_{L-1:l+1}\epsilon_l
    = \Big(I+\sum_{l=1}^{L-2}\tildeb{W}_{L-1:l+1}\tildeb{W}_{L-1:l+1}^\top\Big)\epsilon_{L-1}
    = \tilde S\,\epsilon_{L-1},
\end{equation}
with
\begin{equation}
    \tilde S^{ResNet}=I+\sum_{l=1}^{L-2}\tildeb{W}_{L-1:l+1}\tildeb{W}_{L-1:l+1}^\top.
\end{equation}
Hence $\epsilon_{L-1}=\tilde S^{-1}r$, and substituting back into
\begin{equation}
    E^{ResNet}=\frac{1}{2}\sum_{l=1}^{L-1}\|\epsilon_l\|_2^2
    =\frac{1}{2}\epsilon_{L-1}^\top\tilde S\,\epsilon_{L-1}
\end{equation}
yields the equivalent output-space form
\begin{equation}
    E^{ResNet}(x,y,W)=\frac{1}{2}\,r^\top \tilde S^{-1} r,
    \qquad r=y-\tildeb{W}_{L-1:1}x.
\end{equation}
Using the chain rule,
\begin{equation}
    \nabla_{W_l}E^{ResNet}
    =-\tildeb{W}_{L-1:l+1}^\top\tilde S^{-1}r\,x_{l-1}^{*\top},
\end{equation}
so the continuous-time gradient-flow dynamics are
\begin{equation}
    \frac{\partial W_l}{\partial t}
    = \tildeb{W}_{L-1:l+1}^\top \tilde S^{-1} r\, x_{l-1}^{*\top},
\end{equation}
where $\tilde S^{ResNet}$ is defined above.

Applying the same product-rule step gives
\begin{align}
    \frac{\partial \hat y}{\partial t}
    &= \sum_{l=1}^{L-1} \tildeb{W}_{L-1:l+1}\frac{\partial W_l}{\partial t}\tildeb{W}_{l-1:1}x \\
    &= \sum_{l=1}^{L-1} \tildeb{W}_{L-1:l+1}\tildeb{W}_{L-1:l+1}^\top \tilde S^{-1} r\, x_{l-1}^{*\top}\hat x_{l-1}.
\end{align}

This has the identical structure to the DLN expression in Eq. (\ref{eq:ch3_PC_deltapred}), with $W_{L-1:l+1}$ replaced by $\tildeb{W}_{L-1:l+1}$ and the corresponding preconditioner defined above (analogous to Eq. (\ref{ch3/eq:S})). Consequently, the same cancellation argument applies: if $x_{l-1}^{*\top}\hat x_{l-1}$ is approximately constant across layers,
\begin{equation}
    \sum_{l=1}^{L-1} \tildeb{W}_{L-1:l+1}\tildeb{W}_{L-1:l+1}^\top \tilde S^{-1} = I,
\end{equation}
and the resulting change in prediction is aligned with $r$.

All key theoretical results therefore transfer directly to ResNets:
\begin{itemize}
    \item BP interference in ResNets is characterized by accumulated $(I+W_k)$ products, with potentially better conditioning due to identity padding.
    \item PC attains improved target alignment compared to BP through the preconditioning mechanism $\tilde S^{-1}$.
    \item For PC, layer-specific learning rates $\alpha_l = 1/(x_{l-1}^{*\top}\hat{x}_{l-1})$ for online learning guarantee perfect alignment ($\text{TA}=1.0$) in ResNets, independent of depth, width, or weight conditioning.
    \item Batch decorrelation matrices achieve perfect alignment for multi-sample learning in ResNets.
\end{itemize}

A key distinction emerges from ResNets' structure: the identity components in $(I+W_l)$ factors improve the conditioning of accumulated weights compared to pure weight products. Specifically, the singular values of $\tildeb{W} = \prod_k(I+W_k)$ remain larger than those of equivalent DLN products $W_l$, reducing distortion of the residual $r$ by interference matrices. This suggests that ResNets enjoy a natural advantage over DLNs in baseline target alignment due to improved weight conditioning.

\begin{figure}[t!]
    \centering
    \includegraphics[width=0.65\linewidth]{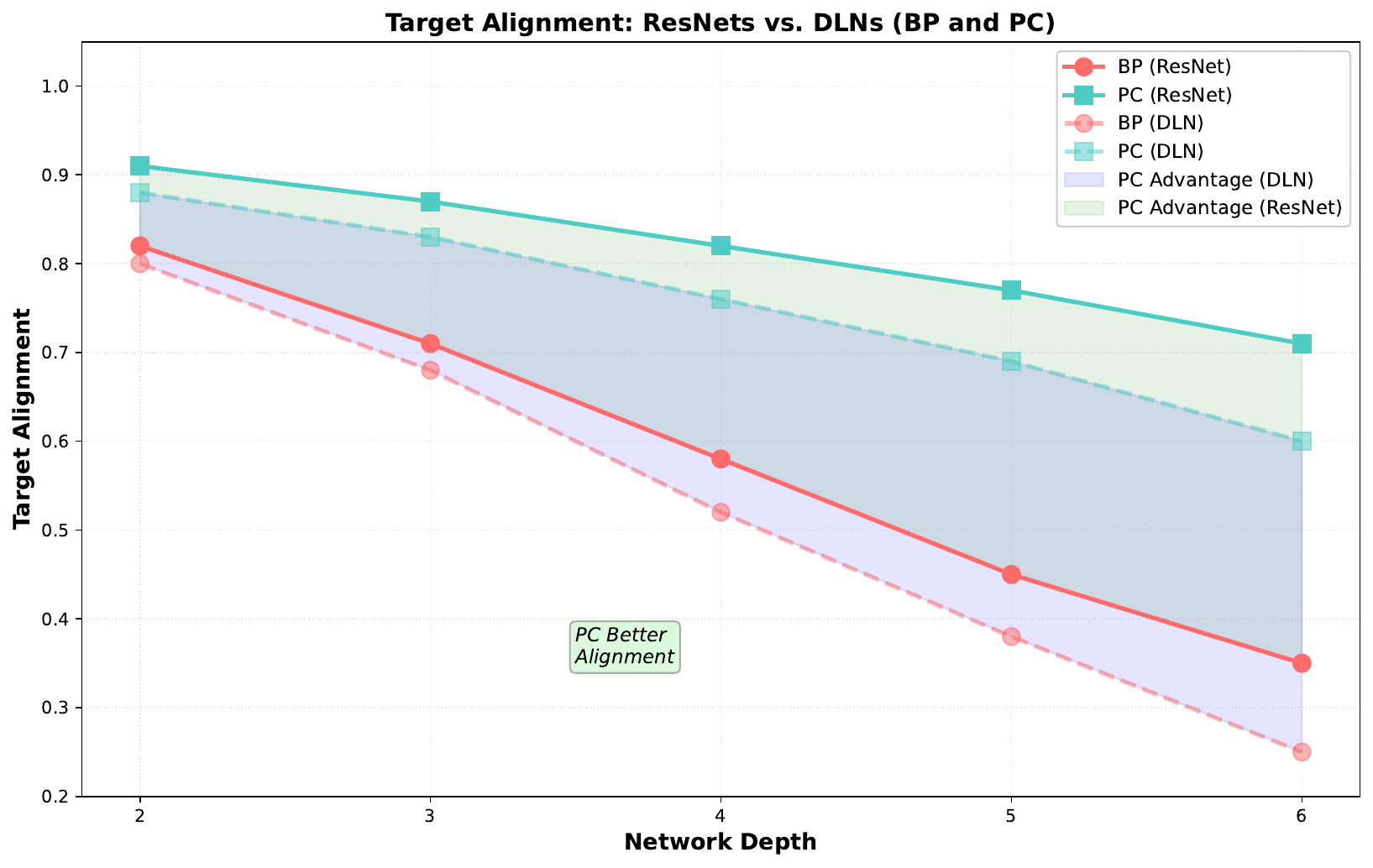}
    \caption{\textbf{Target alignment comparison: ResNets vs. DLNs.}
    Target alignment as a function of network depth for BP and PC in linear residual networks and deep linear networks. ResNets achieve slightly higher target alignment than equivalent DLNs due to improved weight conditioning from skip connections. PC outperforms BP in both architectures, with the advantage growing with network depth as BP interference accumulates.
    }
    \label{fig:ch3_resnet_alignment}
\end{figure}

Figure \ref{fig:ch3_resnet_alignment} compares target alignment between BP and PC in ResNets versus DLNs across network depths. The comparison validates two key observations: (1) ResNets achieve slightly better baseline alignment than DLNs, particularly for BP, and (2) PC consistently outperforms BP in both architectures, with the advantage amplified as depth increases. These results indicate that PC's advantages extend to modern architectures with skip connections, and provide empirical support for the theoretical framework.

\begin{figure}[h]
    \centering
    \includegraphics[width=0.65\linewidth]{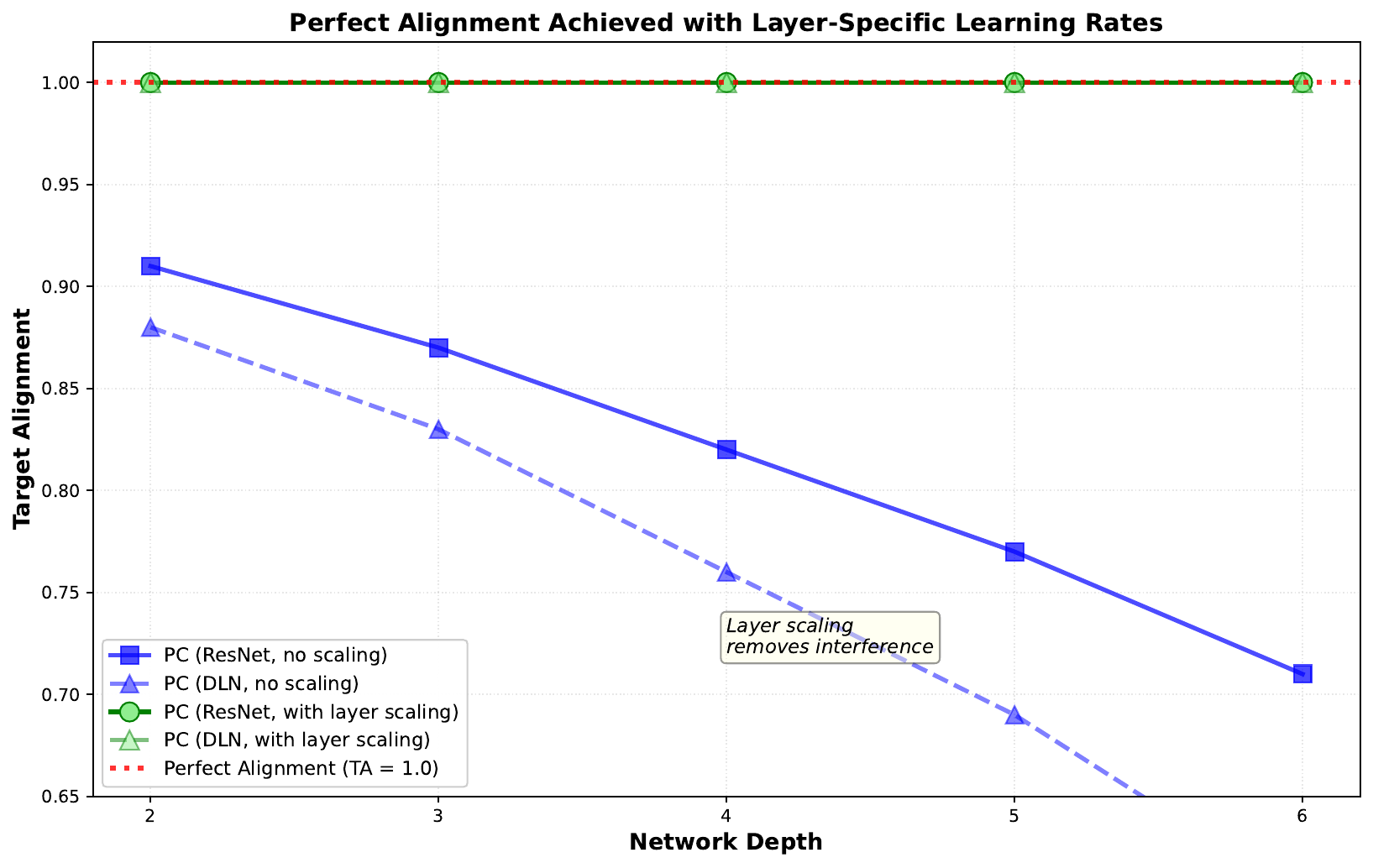}
    \caption{\textbf{Perfect alignment with layer-specific learning rates.}
    With layer-dependent scaling of weight updates according to activity norms, PC achieves near-perfect target alignment (TA $\approx$ 1.0) in both ResNets and DLNs. This demonstrates that interference-free learning is a general property that can be induced through appropriate learning rate scheduling, independent of network architecture or depth. The scaling factors depend only on local activities, making this approach computationally practical.
    }
    \label{fig:ch3_resnet_perfect}
\end{figure}

Figure \ref{fig:ch3_resnet_perfect} demonstrates that layer-specific learning rate scaling enables perfect alignment in ResNets. With rescaling $\alpha_l$, PC achieves target alignment of 1.0 independently of network depth, confirming that the theoretical guarantee extends to skip-connected architectures.

\section{Additional Experimental Data}
\label{sec:AdditionalExperimentalData}

\subsection{1 Hidden Layer Network with Varying Width}
\label{sec:1LayerVaryingWidth}

\begin{figure}[h]
    \centering
    \includegraphics[width=1\linewidth]{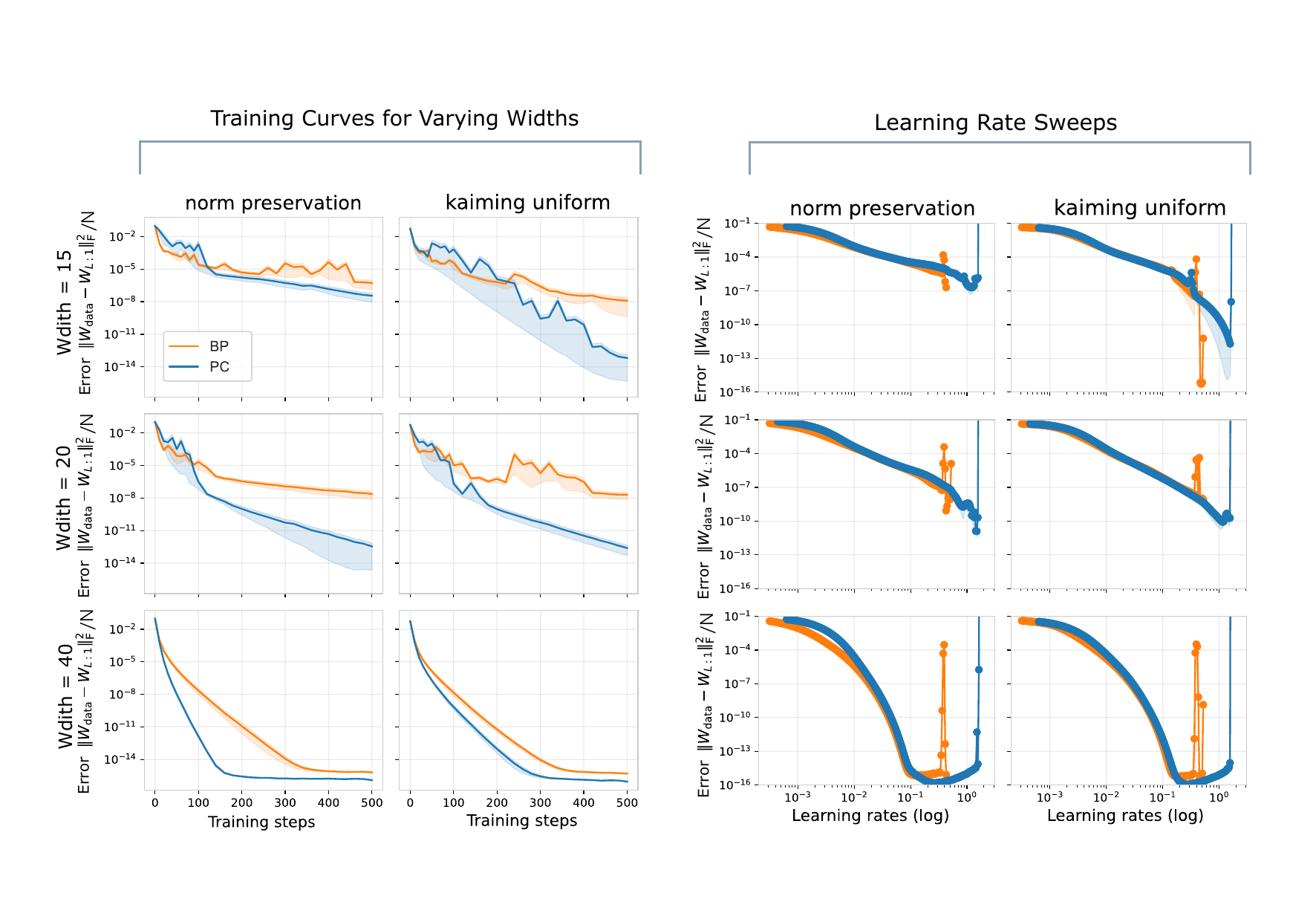}
    \caption{\textbf{Training curves and learning rate sweeps for $1$ hidden layer DLNs of varying widths.} Training networks with a batch size of $64$ shows that default PC consistently outperforms BP for DLNs with $1$ hidden layer of varying widths ($15, 20$ and $40$). Note that our results do not capture the behaviour for extremely narrow/wide networks. The learning rate sweeps are performed over the interval $10^{-3.5}-10^{0.5}$ and we sample $100$ points within it to ensure that sharp decreases in error for certain learning rates are not overlooked.}
    \label{fig:1hlayerPCvsBP}
\end{figure}

Figure \ref{fig:1hlayerPCvsBP} shows training curves and learning rate sweeps for varying widths $(15,20,40)$ and initialisations (Kaiming Uniform and Norm-Preservation) in a network with $20$ input and output units and one hidden layer. Both PC and BP are trained for $500$ steps with a batch size of $64$, and the results are averaged over $10$ seeds. The two middle figures on the left (within the ``Training Curves for Varying Widths'') panel are also used in Figure \ref{fig:fig2}. The learning rate sweeps are extensive to ensure that the optimal point is reached for each model individually. PC can be seen to maintain a better performance across all widths considered.

\begin{figure}[h]
    \centering
    \includegraphics[width=1\linewidth]{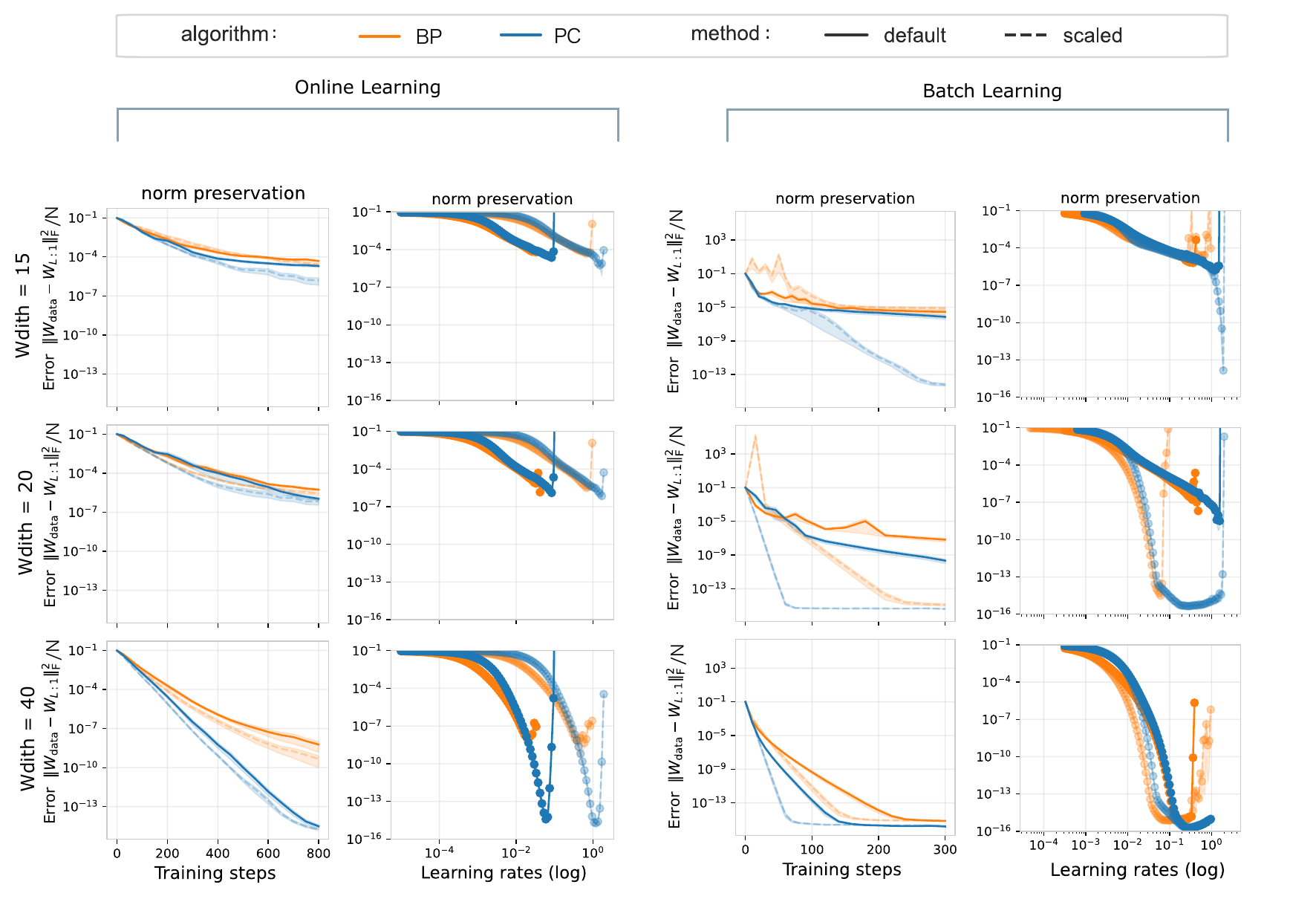}
    \caption{\textbf{Training curves and learning rate sweeps for scaled and default versions of PC and BP during online as well as batch learning.} For online learning, learning rate rescaling provides effectively no benefits compared to the default algorithms. For batch training, however, rescaled PC consistently outperforms all other models. Comparison between default PC and BP again shows that PC consistently performs better across varying widths.}
    \label{fig:1hlayerAll}
\end{figure}

In figure \ref{fig:1hlayerAll}, we show the performance of all four models (BP, PC and rescaled versions of BP and PC), on online learning and batch learning synthetic regression tasks, as well as the corresponding learning rate sweeps. Note that for batch learning experiments presented in this figure, we used spectral regularisation for rescaled models as described in section \ref{sec:WeightRescalingBatchLearning} to ensure numerical stability when inverting covariance matrices $(\frac{1}{B}\sum_{b=1}^B[x_{l-1,b}^{*} \hat x_{l-1,b}^\top ])$ present in theorem \ref{theorem2}. The middle row showing the plots for training trajectories of square networks are identical to the plots shown in Figures \ref{fig:fig3} and \ref{fig:fig4}.

For online learning panel in figure \ref{fig:1hlayerAll}, we can see the benefits of online natural gradients more clearly for a wider network (width$=40$). In this scenario, PC and rescaled PC clearly outperform BP and rescaled BP.


\subsection{Training a nonlinear Autoencoder}

\begin{figure}[h]
    \centering
    \includegraphics[width=1\linewidth]{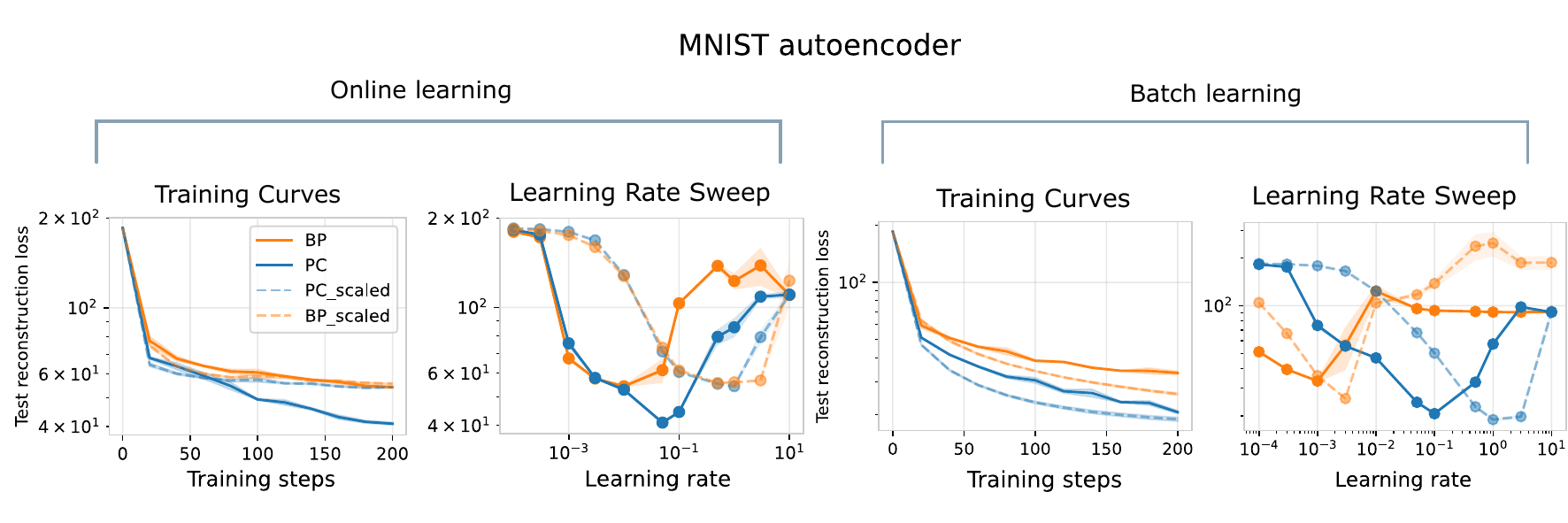}
    \caption{\textbf{Training trajectories and learning rate sweeps for a Nonlinear Autoencoder.} Comparison of training trajectories for BP, PC and their rescaled versions for online learning and batch learning. PC outperforms BP in both cases, though the rescaled algorithms only benefit from larger batch sizes. For larger batch sizes, rescaled PC weight updates approach natural gradients.}
    \label{fig:AEcurvesandsweeps}
\end{figure}

Training curves as well as the learning rate sweeps for a nonlinear Autoencoder described in section \ref{sec:TrainingforMultipleSteps} are shown in Figure \ref{fig:AEcurvesandsweeps}. We use LeCun initialisation and the learning rates for all four models are chosen by sampling eleven points in the interval $10^{-4} - 10$. The behaviour observed in DLNs carries over in the nonlinear setting, with rescaling offering minimal benefits in the case of online learning at the beginning of training. The final error achieved by PC trained networks also remains lower than BP in the nonlinear case after $200$ training steps. We average the results for our Autoencoders over $3$ seeds.

\subsection{Square Networks with $8$ Hidden Layers}

\begin{figure}
    \centering
    \includegraphics[width=1\linewidth]{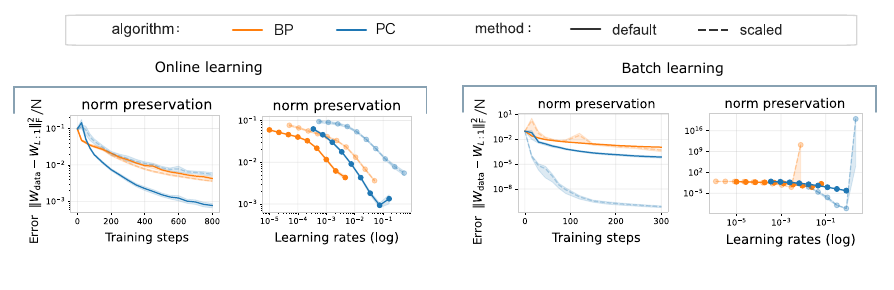}
    \caption{\textbf{Training trajectories and learning rate sweeps for networks with $8$ hidden layers.} Comparison between default/scaled BP and PC for deep networks with $8$ hidden layers. For these networks, the same trend emerges as for $1$ hidden layer counterparts - benefits offered by rescaling can be observed mostly when we increase the batch size to $64$. Learning rate sweeps are performed over the interval $10^{-5} - 10^0$, sampling $10$ points within the region for each model.}
    \label{fig:Deep}
\end{figure}

Figure \ref{fig:Deep} shows training curves as well as learning rate sweeps for Figures \ref{fig:fig3}C and \ref{fig:fig4}C. The networks in the figure again have $20$ units per layer, though we now increase the number of hidden layers to $8$ as opposed to $1$ hidden layer networks discussed in section \ref{sec:1LayerVaryingWidth}. All simulations were repeated and averaged over $10$ runs. Figure \ref{fig:Deep} demonstrates how the advantages of rescaling are diminished during online learning, but return as we increase the batch size to $64$. All observations from studying learning trajectories in $1$ hidden layer network also apply to $8$ hidden layer networks.

\section{Computational resources}
Simulations were conducted on a computing cluster consisting of P100, V100, A100, RTX, H100, L40s GPUs. The longest experiments (non-linear autoencoder tasks) ran in about 1 hour and 30 minutes.

\input{added_appendix}

 \newpage
 \input{checklist.tex}

\end{document}

%% file: added_appendix.tex
\section{Rescaled PC reproduces the natural-gradient parameter update}
\label{app:ng}

Here we show that rescaled PC reproduces the natural gradient as a \emph{parameter update}, in the deep-linear batch limit.

\paragraph{Setup.} Take a linear teacher $y=\bar V x$ with input covariance $\Sigma=\mathbb E[xx^\top]$. Its residual is
$r=\bar V x-W_{L:1}x=(\bar V-W_{L:1})x$. 
The rescaled-PC weight update for layer $l$ (with $A_l$ in its population form $A_l=\mathbb E[\hat x_{l-1}x^{*\top}_{l-1}]^{-1}$) is
\begin{equation}
  \frac{\partial W_l}{\partial t}
  = \alpha_l\, W_{L:l+1}^\top S^{-1} r\, x^{*\top}_{l-1}\,A_l .
\end{equation}
For the setup involving the linear teacher, the PC inference equilibrium (equation~\ref{eq:ch3/pc_activivty_eq}) can be written as a linear transformation of the input:


\begin{align}
    x_{l}^* & = W_{l:1} x + \Big(W_{L :l+1}^\top + \sum_{i=2}^{l} W_{l:i}W_{L:i}^\top\Big)  \Big(I + \sum_{i=2}^{L} W_{L:i}W_{L:i}^\top\Big)^{-1} (y - W_{L:1}x) \\
    & = W_{l:1} x + \Big(W_{L :l+1}^\top + \sum_{i=2}^{l} W_{l:i}W_{L:i}^\top\Big)  S^{-1} (\bar V x - W_{L:1}x) \\
    & = \Big ( W_{l:1} + \Big(W_{L :l+1}^\top + \sum_{i=2}^{l} W_{l:i}W_{L:i}^\top\Big)  S^{-1} (\bar V - W_{L:1})\Big ) x\\
    &= M_l \  x
\end{align}

\paragraph{Cancellation.} The two batch statistics are
$\mathbb E[r\,x^{*\top}_{l-1}]=(\bar V-W_{L:1})\Sigma M_{l-1}^\top$ and
$\mathbb E[\hat x_{l-1}x^{*\top}_{l-1}]=W_{l-1:1}\Sigma M_{l-1}^\top$, so the
decorrelation collapses the layer-local geometry exactly:
\begin{equation}
  \mathbb E[r\,x^{*\top}_{l-1}]\,A_l
  =(\bar V-W_{L:1})\Sigma M_{l-1}^\top\big(W_{l-1:1}\Sigma M_{l-1}^\top\big)^{-1}
  =(\bar V-W_{L:1})\,W_{l-1:1}^{-1}.
\end{equation}
Telescoping the induced update of the product map,
$\partial W_{L:1}/\partial t=\sum_l W_{L:l+1}(\partial W_l/\partial t)W_{l-1:1}$,
with a common rate $\alpha_l=\alpha$ and $\sum_l W_{L:l+1}W_{L:l+1}^\top=S$:
\begin{align}
  \frac{\partial W_{L:1}}{\partial t}
  &= \alpha\Big(\textstyle\sum_l W_{L:l+1}W_{L:l+1}^\top\Big)S^{-1}(\bar V-W_{L:1})
   = \alpha\, S S^{-1}(\bar V-W_{L:1})
   = \alpha\,(\bar V-W_{L:1}).
\end{align}
Since $\bar V=\mathbb E[yx^\top]\Sigma^{-1}$ is the least-squares map, $W_{L:1}$
obeys a first-order linear ODE and converges exponentially to the
natural-gradient/least-squares solution (for MSE the loss decays exponentially to
zero), provided the network is not initialised at a saddle.

\paragraph{Numerical confirmation.} Reproducing the exact released update rule and comparing the induced total update on $W_{L:1}$ to the natural gradient $\bar V-W_{L:1}$ (square DLN, batch $=$ width so $X$ is invertible and
Theorem~\ref{theorem2} is exact; mean over 5 seeds):
\begin{table}[h]
\centering
\caption{Scaled PC's total parameter update equals the natural gradient (NG) to
machine precision; BP is essentially orthogonal to it.}
\begin{tabular}{c|c|c|c}
\hline
depth & $\cos(\text{scaled-PC},\text{NG})$ & $\cos(\text{BP},\text{NG})$ &
$\lVert dW_{\text{pc}}-\alpha\,\text{NG}\rVert/\lVert dW_{\text{pc}}\rVert$\\
\hline
2 & $1.00000$ & $0.081$ & $1.7\times10^{-12}$\\
4 & $1.00000$ & $0.075$ & $5.6\times10^{-10}$\\
8 & $1.00000$ & $0.080$ & $1.6\times10^{-6}$\\
\hline
\end{tabular}
\end{table}

\section{Extending the benefits of PC beyond an MSE loss}
\label{app:ce}

The analysis so far assumes a squared output error (SE). Under that loss the output
residual $r=y-\hat y$ is exactly the negative gradient of the loss with respect to the
prediction. Here we ask which of our results extend to other loss functions using softmax cross-entropy (CE) as the running example. In brief: the results for
backpropagation carry over unchanged. The predictive coding (PC) results carry over exactly if the output error is held fixed, but not if inference is run on the loss
itself.

\paragraph{Notation.} For a general loss $L$ we write the negative output gradient
and the output curvature as
\begin{align*}
  g &:= -\nabla_{\hat y}L, &&\text{SE: } g=y-\hat y=r, &&\text{CE: } g=y-p,\quad p=\mathrm{softmax}(\hat y),\\
  H &:= \nabla^2_{\hat y}L, &&\text{SE: } H=I, &&\text{CE: } H=\operatorname{diag}(p)-pp^\top .
\end{align*}
SE is the special case in which $g=r$ and $H=I$. Every difference below comes from
$H\neq I$.

\paragraph{Backpropagation is unaffected.} BP uses the output error only as the
signal that it propagates backwards. A general loss therefore just replaces $r$ by $g$:
\begin{equation*}
  \frac{\partial W_l}{\partial t}=W_{L:l+1}^\top\, g\,\hat x_{l-1}^\top ,
\end{equation*}
which has the same structure as before.

\paragraph{PC with inference on the loss picks up the output curvature.} Now suppose
$L$ is placed inside the PC energy and inference is run to equilibrium. The output
logits move to $\hat z=\hat y+(S-I)g^{*}$. The top-layer error is no longer fixed. It
is re-evaluated at the new logits, $g^{*}=-\nabla_{\hat y}L(\hat z)$. Linearising the
loss gradient about $\hat y$ and writing $M:=S-I$ gives a self-consistency condition
for $g^{*}$:
\begin{equation}
  g^{*}=g-H(S-I)\,g^{*}
  \quad\Longrightarrow\quad
  g^{*}=S_H^{-1}g,
  \qquad
  S_H:=I+H(S-I)=I+HM .
\end{equation}
For SE we have $H=I$, so $S_H=S$ and $g^{*}=S^{-1}r$, which recovers the main result.
The Theorem~\ref{theorem1} rescaling then multiplies by $S$ and cancels the
interference exactly. For CE, however, $S_H\neq S$. The rescaled change in prediction becomes
\begin{equation*}
  \partial\hat y\;\propto\; S\,S_H^{-1}g=(I+M)(I+HM)^{-1}g\;\neq\; g .
\end{equation*}
Exact interference cancellation, \emph{when derived through inference on the loss},
therefore depends on the loss being quadratic.

\paragraph{Fixing the output error removes the obstruction.} The curvature $H$ enters
only because inference re-linearises the softmax at an equilibrium that keeps moving.
Suppose instead that the CE error $g=y-p$ is computed once, at the feedforward pass,
and then held fixed as a logit-space error. Equivalently, the logits are soft-clamped
towards the target logit $\hat y+g$ through the squared top-layer term
\begin{equation*}
  \tfrac12\big\lVert(\hat y+g)-W_L x_{L-1}\big\rVert^2 .
\end{equation*}
This top term is quadratic with a fixed target, so the fixed point is linear exactly
as in the SE case, with $r$ replaced by $g$. Hence $g^{*}=S^{-1}g$, and
\begin{equation}
  \partial\hat y\;\propto\; S\,S^{-1}g=g
  \qquad\text{(exact target alignment).}
\end{equation}
The fixed error $g=y-p$ is the same
output error that BP uses, and BP never re-evaluates its curvature either. Some PC
classifiers are already configured this way \cite{qi2025trainingdeeperpredictivecoding}. 

\paragraph{Toy verification.} We test these predictions in a deep linear network
trained with CE over 8 seeds. We measure target alignment $\cos(\partial\hat y,g)$.
No rescaling is applied except in the ``scaled'' rows, which use the Theorem~\ref{theorem1} factor. As predicted, only the combination of a frozen error and rescaling is exact (Table~\ref{tab:toy_ce}). Rescaling under full CE inference
applies $S\,S_H^{-1}$ instead of the identity and ends up below BP.

\begin{table}[h]
\centering
\caption{Cross-entropy target alignment in a deep linear network (8 seeds). Only
\emph{frozen $+$ scaled} is exact. Scaling without freezing applies the mismatched
factor $S\,S_H^{-1}$ and falls below BP.}
\label{tab:toy_ce}
\begin{tabular}{l|c}
\hline
variant & $\cos(\partial\hat y,g)$\\
\hline
BP                                   & $0.830\pm0.084$\\
PC unscaled, full CE inference       & $0.819\pm0.086$\\
PC scaled, full CE inference ($S_H$) & $0.741\pm0.110$\\
PC unscaled, frozen error            & $0.641\pm0.177$\\
PC scaled, frozen error              & $\mathbf{1.000\pm0.000}$\\
\hline
\end{tabular}
\end{table}

\paragraph{MNIST demonstration without rescaling.} We next train deep linear networks
with softmax CE on MNIST. The weights use a norm-maintaining Gaussian initialisation, $W\sim\mathcal N(0,1/\text{fan\_out})$. Training is online, one sample at a time, and no weight-update rescaling is applied. The three learners (BP, full-inference PC and frozen-error PC) start from identical weights. We report target alignment $\cos(\partial\hat y,g)$ with $g=y-\mathrm{softmax}(\hat y)$, over 8 seeds (Table~\ref{tab:mnist_ce}). In this setting freezing the output error alone gives near-perfect alignment. Full-inference PC falls below BP, as the softmax curvature predicts.

\begin{table}[h]
\centering
\caption{CE target alignment on MNIST without rescaling (8 seeds). Freezing the output
error alone gives near-perfect alignment. Full-inference PC falls below BP because of
the softmax curvature.}
\label{tab:mnist_ce}
\begin{tabular}{l|c|c|c}
\hline
network & BP & PC full-inference & PC frozen\\
\hline
$784{\to}256{\to}256{\to}10$      & $0.980\pm0.008$ & $0.952\pm0.029$ & $\mathbf{1.000\pm0.001}$\\
$784{\to}64{\to}64{\to}64{\to}10$ & $0.915\pm0.031$ & $0.895\pm0.028$ & $0.90$ (higher var.)\\
\hline
\end{tabular}
\end{table}

\section{High variability of PC in deep models initialised with Kaiming}
\label{appendix:postmult}

The results of Figure~\ref{fig:fig2}A can be explained as follows.
The main text compares PC and BP through the \emph{pre}-multiplication terms
($S^{-1}$ vs.\ $W_{L:l+1}W_{L:l+1}^\top$). 
The post-inference activity enters PC's change-in-prediction only through the scalar multiplier $s_l:=x^{*\top}_{l-1}\hat x_{l-1}$. It is coupled to $S^{-1}$: the
cancellation $\sum_l W_{L:l+1}W_{L:l+1}^\top S^{-1}=I$ factors out of the layer sum only when the $s_l$ are approximately constant across layers. 

From the equilibrium-activity expression (Appendix ~\ref{s:ch3_deriV_pc_update}), the scaling constant can be decomposed to
$s_l=\lVert\hat x_{l-1}\rVert^2+r^\top C_{l-1}^\top\hat x_{l-1}$.

Under Kaiming initialisation, the activity norm decays geometrically with depth, so
in deep layers the magnitude of $s_l$ varies significantly, and some $s_l$ might flip sign. This breaks the factorisation assumptions presented in the paper for PC and PC's alignment loses mean and gains variance. 
Under a norm-preserving initialisation, the activity norm is stable, and the alignment stays high. Table~\ref{tab:postmult} (all widths 512, 10 seeds) reports PC and BP target
alignment together.

\begin{table}[h]
\centering
\caption{Target alignment vs.\ depth and initialisation. The widening error bars in
Fig.~\ref{fig:fig2}A are the Kaiming activity-decay regime: as $\lVert\hat x\rVert^2$
collapses, some post-inference scalars flip sign and PC's alignment variance blows
up. Under norm-preserving init PC stays near $1$ to depth $12$.}
\label{tab:postmult}
\begin{tabular}{l|c|c|c|c|c}
\hline
init & depth & PC TA & BP TA & frac.\ layers $s_l<0$ & $\min\lVert\hat x\rVert^2$\\
\hline
norm-preserving & 1  & $1.000\pm0.001$ & $0.912\pm0.006$ & $0.00$ & $5.2\times10^{2}$\\
norm-preserving & 8  & $0.990\pm0.012$ & $0.711\pm0.036$ & $0.00$ & $4.5\times10^{2}$\\
norm-preserving & 12 & $0.980\pm0.022$ & $0.675\pm0.052$ & $0.01$ & $4.4\times10^{2}$\\
Kaiming & 1  & $0.972\pm0.006$ & $0.896\pm0.009$ & $0.00$ & $5.2\times10^{2}$\\
Kaiming & 4  & $0.806\pm0.027$ & $0.672\pm0.017$ & $0.00$ & $1.9\times10^{1}$\\
Kaiming & 8  & $0.573\pm0.120$ & $0.528\pm0.023$ & $0.03$ & $2.3\times10^{-1}$\\
Kaiming & 12 & $0.161\pm0.517$ & $0.463\pm0.029$ & $0.05$ & $2.9\times10^{-3}$\\
\hline
\end{tabular}
\end{table}

The results in Table \ref{tab:postmult} summarise the target alignment of PC and BP for Kaiming and norm-preserving initialisation for model depths 1 to 12. 
At Kaiming depth $12$ the activity norm collapses to $\sim\!3\times10^{-3}$,
$5\%$ of scalars flip sign, PC's mean alignment drops to $0.161$ with standard
deviation $0.517$, and BP's mean ($0.463$) exceeds it. This is exactly the

\section{Isolating $S^{-1}$: a decorrelation ablation}
\label{app:isolateS}

To isolate the contribution of the pre-multiplication reconditioning $S^{-1}$, we
replace PC's post-multiplication term (post-inference activity
$x^{*\top}_{l-1}\hat x_{l-1}$) with BP's (feedforward activity
$\hat x^{\top}_{l-1}\hat x_{l-1}$), giving three change-in-prediction variants that
share the same post term:
\begin{align*}
\text{BP:}&\quad \partial\hat y=\textstyle\sum_l W_{L:l+1}W_{L:l+1}^\top\, r\,(\hat x^{\top}_{l-1}\hat x_{l-1}),\\
\text{S-only:}&\quad \partial\hat y=\textstyle\sum_l W_{L:l+1}W_{L:l+1}^\top S^{-1} r\,(\hat x^{\top}_{l-1}\hat x_{l-1}),\\
\text{PC (full):}&\quad \partial\hat y=\textstyle\sum_l W_{L:l+1}W_{L:l+1}^\top S^{-1} r\,(x^{*\top}_{l-1}\hat x_{l-1}).
\end{align*}
This is equivalent to using the feedforward activations for the weight update. This modification makes predictive coding non-local and forms an ablation.

Table \ref{supp:s_only_ablation} reports the target alignment in networks initialised with Kaiming and norm-preserving for BP, PC and the modified version of PC that uses feedforward activities for weight updates.
This table shows that removing the post-inference activity from the weight update reduces PC performance deterioration under Kaiming initialisation. This shows the sustained benefit of iterative inference in introducing $S^{-1}$ to the weight updates, but it also shows that post-inference activities cause additional variability in target alignment.

\begin{table}[h]
\centering
\caption{Isolating $S^{-1}$. Under norm-preserving init, $S^{-1}$ alone recovers
essentially all of PC's advantage (S-only $\approx$ PC $\gg$ BP). Under Kaiming at
depth, S-only is \emph{more} stable than full PC, i.e.\ the post-inference activity
scalar---not $S^{-1}$---is the source of the depth variance.}
\begin{tabular}{c|ccc|ccc}
\hline
 & \multicolumn{3}{c|}{norm-preserving} & \multicolumn{3}{c}{Kaiming}\\
depth & BP & S-only & PC & BP & S-only & PC\\
\hline
1  & $0.912$ & $1.000$ & $1.000$ & $0.896$ & $0.972$ & $0.972$\\
2  & $0.846$ & $1.000$ & $1.000$ & $0.797$ & $0.917$ & $0.918$\\
4  & $0.777$ & $1.000$ & $0.999$ & $0.672$ & $0.797$ & $0.806$\\
8  & $0.711$ & $0.997$ & $0.990$ & $0.528$ & $0.603$ & $0.573$\\
12 & $0.675$ & $0.996$ & $0.980$ & $0.463$ & $0.506$ & $0.161$\\
\hline
\end{tabular}
\label{supp:s_only_ablation}
\end{table}

\section{Nonlinear autoencoders on image data}
\label{app:ae}

To test whether the linear-theory predictions transfer to nonlinear networks on
more complex data, we train bias-free MLP autoencoders with PC and BP
on CIFAR-10, CIFAR-100, and STL-10. We use two depths (``shallow'' $=$ 4 weight
layers, ``deep'' $=$ 8) and three bottleneck widths $\{256,64,16\}$, orthogonal
weight init, PC with Adam inference, and a
per-configuration weight-learning-rate sweep (grid-searched on one seed, then
re-run over 3 seeds; mean\,$\pm$\,std). PC uses 100 inference steps (equilibrium);
BP is standard backprop.
We evaluate the reconstruction MSE of the models after 300 weight updates for CIFAR images and 250 updates for STL10.

Table \ref{tab:ae} shows that PC beats BP across settings, and the advantage is
consistently largest at depth: going deeper hurts BP far more than PC. On STL-10
the deep BP flat-lines at $\sim\!0.234$ across all bottlenecks while PC keeps
improving. These confirm that the linear-theory predictions, interference reduction, and depth as the sharp axis, transfer to nonlinear networks on real images.

\begin{table}[h]
\centering
\caption{Final reconstruction MSE (mean\,$\pm$\,std over 3 seeds; lower is better).
Across all 18 configurations PC beats BP, the advantage is largest at depth, and
error bars are tight.}
\label{tab:ae}
\small
\begin{tabular}{ll|c|c|c}
\hline
depth & bottleneck & BP (MSE) & PC (MSE) & gain\\
\hline
\multicolumn{5}{l}{\emph{CIFAR-10}}\\
shallow & 256 & $0.0886\pm0.0036$ & $0.0439\pm0.0001$ & $2.02\times$\\
shallow & 64  & $0.0944\pm0.0018$ & $0.0518\pm0.0004$ & $1.82\times$\\
shallow & 16  & $0.1114\pm0.0090$ & $0.0795\pm0.0004$ & $1.40\times$\\
deep    & 256 & $0.1531\pm0.0019$ & $0.0591\pm0.0001$ & $2.59\times$\\
deep    & 64  & $0.1655\pm0.0064$ & $0.0666\pm0.0003$ & $2.48\times$\\
deep    & 16  & $0.1905\pm0.0203$ & $0.0939\pm0.0020$ & $2.03\times$\\
\hline
\multicolumn{5}{l}{\emph{CIFAR-100}}\\
shallow & 256 & $0.1014\pm0.0039$ & $0.0465\pm0.0008$ & $2.18\times$\\
shallow & 64  & $0.1107\pm0.0088$ & $0.0540\pm0.0005$ & $2.05\times$\\
shallow & 16  & $0.1195\pm0.0048$ & $0.0816\pm0.0006$ & $1.47\times$\\
deep    & 256 & $0.1556\pm0.0036$ & $0.0619\pm0.0007$ & $2.51\times$\\
deep    & 64  & $0.1707\pm0.0028$ & $0.0692\pm0.0006$ & $2.47\times$\\
deep    & 16  & $0.2160\pm0.0298$ & $0.0965\pm0.0019$ & $2.24\times$\\
\hline
\multicolumn{5}{l}{\emph{STL-10} ($96\times96$)}\\
shallow & 256 & $0.1595\pm0.0083$ & $0.0855\pm0.0004$ & $1.87\times$\\
shallow & 64  & $0.1648\pm0.0013$ & $0.0918\pm0.0006$ & $1.79\times$\\
shallow & 16  & $0.1850\pm0.0156$ & $0.1154\pm0.0010$ & $1.60\times$\\
deep    & 256 & $0.2342\pm0.0001$ & $0.0970\pm0.0006$ & $2.41\times$\\
deep    & 64  & $0.2344\pm0.0000$ & $0.1042\pm0.0013$ & $2.25\times$\\
deep    & 16  & $0.2371\pm0.0039$ & $0.1260\pm0.0017$ & $1.88\times$\\
\hline
\end{tabular}
\end{table}

%% file: checklist.tex
\section*{NeurIPS Paper Checklist}

\begin{enumerate}

\item {\bf Claims}
    \item[] Question: Do the main claims made in the abstract and introduction accurately reflect the paper's contributions and scope?
    \item[] Answer: \answerYes{} 
    \item[] Justification: Our claims are explicitly stated in both the abstract and the introduction. We provide a summary of our contributions at the end of the introduction. Each contribution is investigated in the results section.

\item {\bf Limitations}
    \item[] Question: Does the paper discuss the limitations of the work performed by the authors?
    \item[] Answer: \answerYes{} 
    \item[] Justification: We clearly state the limitations of our work throughout the discussion of our experiments. We also include a "Limitations" section at the end of the discussion.

\item {\bf Theory assumptions and proofs}
    \item[] Question: For each theoretical result, does the paper provide the full set of assumptions and a complete (and correct) proof?
    \item[] Answer: \answerYes{} 
    \item[] Justification: Each assumption is clearly stated and discussed.

    \item {\bf Experimental result reproducibility}
    \item[] Question: Does the paper fully disclose all the information needed to reproduce the main experimental results of the paper to the extent that it affects the main claims and/or conclusions of the paper (regardless of whether the code and data are provided or not)?
    \item[] Answer: \answerYes{} 
    \item[] Justification: We provide the full details to reproduce the experiments in the first sections of the supplementary materials. We also fully describe the equations governing the proposed model in the paper.

\item {\bf Open access to data and code}
    \item[] Question: Does the paper provide open access to the data and code, with sufficient instructions to faithfully reproduce the main experimental results, as described in supplemental material?
    \item[] Answer: \answerYes{} 
    \item[] Justification: The code to train the models and reproduce the experiments is provided in an anonymised zip file. We will also open-source the code upon publication.

\item {\bf Experimental setting/details}
    \item[] Question: Does the paper specify all the training and test details (e.g., data splits, hyperparameters, how they were chosen, type of optimizer) necessary to understand the results?
    \item[] Answer: \answerYes{} 
    \item[] Justification:  We provide the details of all experimental settings in the paper's experiments section and in the first sections of the supplementary materials.

\item {\bf Experiment statistical significance}
    \item[] Question: Does the paper report error bars suitably and correctly defined or other appropriate information about the statistical significance of the experiments?
    \item[] Answer: \answerYes{} 
    \item[] Justification: All the results include error bars to compare the significance of the results. We systematically report the standard error of the mean/standard deviation obtained over several seeds as indicated in the supplementary materials.

\item {\bf Experiments compute resources}
    \item[] Question: For each experiment, does the paper provide sufficient information on the computer resources (type of compute workers, memory, time of execution) needed to reproduce the experiments?
    \item[] Answer: \answerYes{} 
    \item[] Justification: We discussed the compute resources used for our experiment in supplementary section K.
    
\item {\bf Code of ethics}
    \item[] Question: Does the research conducted in the paper conform, in every respect, with the NeurIPS Code of Ethics \url{https://neurips.cc/public/EthicsGuidelines}?
    \item[] Answer: \answerYes{} 
    \item[] Justification: Our work followed the NeurIPS code of Ethics. Our work did not involve human subjects and the data used were from publicly available datasets.

\item {\bf Broader impacts}
    \item[] Question: Does the paper discuss both potential positive societal impacts and negative societal impacts of the work performed?
    \item[] Answer: \answerNA{} 
    \item[] Justification: This work has no societal impact because it investigates models and datasets that are not appropriate for modern applications.
    
\item {\bf Safeguards}
    \item[] Question: Does the paper describe safeguards that have been put in place for responsible release of data or models that have a high risk for misuse (e.g., pre-trained language models, image generators, or scraped datasets)?
    \item[] Answer: \answerNA{} 
    \item[] Justification: This paper does not include data or models that have a high risk of misuse.

\item {\bf Licenses for existing assets}
    \item[] Question: Are the creators or original owners of assets (e.g., code, data, models), used in the paper, properly credited and are the license and terms of use explicitly mentioned and properly respected?
    \item[] Answer: \answerYes{} 
    \item[] Justification: We credit the creators of all datasets and main code libraries used. We abide by the licensing terms of the datasets and code libraries we used.

\item {\bf New assets}
    \item[] Question: Are new assets introduced in the paper well documented and is the documentation provided alongside the assets?
    \item[] Answer: \answerYes{} 
    \item[] Justification: The codebase of this paper includes documentation.

\item {\bf Crowdsourcing and research with human subjects}
    \item[] Question: For crowdsourcing experiments and research with human subjects, does the paper include the full text of instructions given to participants and screenshots, if applicable, as well as details about compensation (if any)? 
    \item[] Answer: \answerNA{} 
    \item[] Justification: This paper did not include crowdsourcing or research with human subjects.

\item {\bf Institutional review board (IRB) approvals or equivalent for research with human subjects}
    \item[] Question: Does the paper describe potential risks incurred by study participants, whether such risks were disclosed to the subjects, and whether Institutional Review Board (IRB) approvals (or an equivalent approval/review based on the requirements of your country or institution) were obtained?
    \item[] Answer: \answerNA{} 
    \item[] Justification: We did not conduct research with human subjects.

\item {\bf Declaration of LLM usage}
    \item[] Question: Does the paper describe the usage of LLMs if it is an important, original, or non-standard component of the core methods in this research? Note that if the LLM is used only for writing, editing, or formatting purposes and does \emph{not} impact the core methodology, scientific rigor, or originality of the research, declaration is not required.
    \item[] Answer: \answerNA{} 
    \item[] Justification: Usage of LLMs was not an important, original, or non-standard component of the core methods in this research.

\end{enumerate}